\documentclass[preprint,review,12pt,3p]{elsarticle}

\usepackage{amssymb}
\usepackage{amsmath}
\usepackage{color}
\usepackage{array}
\usepackage{booktabs}
\usepackage{multirow}
\usepackage{xcolor}
\usepackage{url}
\usepackage{algorithmic}
\usepackage{algorithm}
\usepackage{makecell}

\usepackage{stackengine}

\journal{Knowledge-Based Systems}
\usepackage[breaklinks,colorlinks,linkcolor=black,citecolor=black,urlcolor=black]{hyperref}

\begin{document}

\begin{frontmatter}



\title{Multimodal Image Matching based on Frequency-domain Information of Local Energy Response}

%

\author[label1,label2,label3]{Meng Yang}
\ead{20171002458@cug.edu.cn}

\author[label1,label2,label3]{Jun Chen\corref{correspondingauthor}}
\ead{chenjun71983@163.com}

\author[label4]{Wenping Gong}
\ead{wenpinggong@cug.edu.cn}

\author[label1,label2,label3]{Longsheng Wei}
\ead{weilongsheng@163.com}

\author[label5]{Xin Tian}
\ead{xin.tian@whu.edu.cn}

\cortext[correspondingauthor]{Corresponding author.}

\affiliation[label1]{organization={School of Automation, China University of Geosciences},city={Wuhan 430074}, country={China}}

\affiliation[label2]{organization={Hubei Key Laboratory of Advanced Control and Intelligent Automation for Complex Systems},city={Wuhan 430074}, country={China}}

\affiliation[label3]{organization={Engineering Research Center of Intelligent Technology for Geo-Exploration, Ministry of Education},city={Wuhan 430074}, country={China}}

\affiliation[label4]{organization={Faculty of Engineering, China University of Geosciences}, city={Wuhan 430074}, country={China}}

\affiliation[label5]{organization={Electronic Information School, Wuhan University}, city={Wuhan 430072}, country={China}}

\begin{abstract}
	
 Complicated nonlinear intensity differences, nonlinear local geometric distortions, noises and rotation transformation are main challenges in multimodal image matching. In order to solve these problems, we propose a method based on Frequency-domain Information of Local Energy Response called FILER. The core of FILER is the local energy response model based on frequency-domain information, which can overcome the effect of nonlinear intensity differences. To improve the robustness to local nonlinear geometric distortions and noises, we design a new edge structure enhanced feature detector  and convolutional feature weighted descriptor, respectively. In addition, FILER overcomes the sensitivity of the frequency-domain information to the rotation angle and achieves rotation invariance. Extensive experiments multimodal image pairs show that FILER outperforms other state-of-the-art algorithms and has good robustness and universality.
\end{abstract}



\begin{keyword}
	
Multimodal image matching \sep Image matching \sep Nonlinear intensity differences \sep Frequency-domain information \sep Frequency-domain filter

\end{keyword}

\end{frontmatter}


\section{Introduction}
Multimodal images are images from different sensors, or images captured under different imaging conditions \cite{jiang2021review}. Multimodal images contain a wealth of information and a wide range of physical properties due to the fact that different sensors acquire signals on different principles and the same sensor records different physical phenomena under different imaging conditions. The goal of multimodal image matching (MIM) is to construct reliable matches between two multimodal images \cite{zhang2021multimodal}. It aligns, unifies and compensates key information in multimodal images, which greatly improves image data utilization and produces richer scene representations. MIM plays an important role in change detection \cite{cheng2024dmf2net}, pattern recognition \cite{fan2022vlsg}, image fusion \cite{luo2024fidelity,mei2024gtmfuse}, clinical disease diagnosis \cite{huang2024multi}, and so on. 

\textbf{The main challenges of MIM} is: (1) the multimodal images have complicated nonlinear intensity differences (NIDs) and the NIDs in different multimodal images are different. In the past decades, many methods have been proposed to remove the NIDs \cite{zhang2021multimodal,li2019rift,yao2022multi}. But there is a lack of general algorithms for different multimodal images; (2) the multimodal images contain local nonlinear geometric distortion, various noise and low texture problems which greatly degrade the performance of MIM for the applications; (3) the multimodal images may contain rotation transformation, which increases the difficulty of matching based on frequency-domain information. Therefore, it is necessary to find an efficient, general, and robust matching algorithm for MIM.

The multimodal images has complex NIDs. NIDs refer to the phenomenon of nonlinear differences in image intensity or gray under different imaging mechanisms. There are three reasons for NIDs: (1) Differences in the imaging properties of sensors. For example, the synthetic aperture radar (SAR) images contain phase information, while the optical images contain amplitude information \cite{deng2022amplitude}. (2) Radiative transmission errors due to the atmosphere. For example, in multi-temporal images, the spectral emissivity of objects is often distorted due to atmospheric action and illumination conditions \cite{mikolajczyk2001indexing}. (3) Severe appearance differences due to different luminosities, imaging times, spectra, and resolutions, such as cross-temporal and cross-spectral image pairs \cite{jiang2021review}. The gradient-based and intensive-based multimodal image processing methods are sensitive to NIDs \cite{li2019rift,yao2022multi,ye2019fast}, and difficult to extract feature points with high repeatability and accuracy. Therefore, multimodal image processing is a difficult problem in image matching.

Our \textbf{motivation} is to construct a robust model to overcome complex NIDs, and on this basis to design new feature detector and descriptor to overcome noises and nonlinear local geometric distortions and achieve rotation invariance. The NIDs mask and corrupt similar structures of multimodal images, and reduce their correlation, which degrades the matching performance. In order to overcome the NIDs, the algorithm needs to find the common features of multimodal images accurately. The common features are edges and bars, which have specific local phase properties in frequency-domain. Therefore, we extract the frequency-domain information of images \cite{arrospide2013log}. Because the second-order nonlinearity of local energy form can detect many natural edges and contours with complex intensity distribution in images well \cite{rosenthaler1992detection}, we construct local energy response $energy_o$ to improve the similarity between multimodal images. Based on $energy_o$, we propose FILER and design novel feature detector and descriptor, which can overcome the problem of NIDs, local nonlinear geometric distortions, and noises. Because the frequency-domain information of the images is very sensitive to the rotation angle (see Section III-D and Figure~\ref{fig4} for details), the frequency-domain based methods cannot cope with the rotation transformation of the images. However, we designs a method to achieve rotation invariance of FILER well.

More specifically, we first convolve the images with a multi-scale and multi-orientation frequency-domian filter to obtain the wavelet response component $E_{so}$ and $O_{so}$ with $N_s$ scales and $N_o$ orientations. Subsequently, we extract the local energy response $energy_o$ with $N_o$ orientations from $E_{so}$ and $O_{so}$. In feature detection, we obtain the total energy response $ET$ with large-scale edge structures by processing the sum of $energy_o$. However, due to the interference of noises and outliers, the large-scale edge structures of $ET$ are blurred. In order to enhance the blurred edge structures, we design an edge structure enhanced detector, which takes $ET$ as the guidance image and uses spatial distance function and structural guidance function to filter the original image. In feature description, we construct the energy manifold vector field by taking the maximum orientation of $energy_o$, which contains rich texture details, so we build the descriptor based on it. In order to overcome local nonlinear geometric distortions on descriptors, we design a convolutional feature weighting method to deal with $energy_o$. Our descriptor is a log-polar structure, which facilitates rotation invariance. Finally, the putative matches are established according to the similarity of descriptors. In addition, we design a new method to achieve rotation invariance. Comparison experiments show that FILER overcomes the main challenges of MIM, and also has achieved the state-of-the-art performance.

The main contributions and novelties of our FILER are:

(1) The local energy response model is proposed. It is the basis of feature detector and descriptor. The local energy response model can detect the common features of multimodal images and overcome the influence of NIDs. 

(2) The new feature detector and descriptor are designed. The edge structure enhanced feature detector improves the robustness to noises. The log-polar descriptor weighted by convolutional feature improves the robustness to local nonlinear geometric distortions.

(3) In order to overcome the problem that frequency-domain based methods are sensitive to the rotation angle, a new method is designed to achieve rotation invariance of the FILER, which is the main direction method.
\section{Related Work}
In many past researches, multimodal images still need to be manually processed and matched, which is time-consuming. Moreover, it inevitably brings some subjective matching errors \cite{ye2019fast}. Therefore, the search for fast and robust MIM algorithms in applications is urgent. Currently, MIM methods can be classified into template-based, deep learning-based and feature-based methods \cite{zhang2021multimodal,chen2022robust}.
\subsection{Template-based Methods}	
Its basic step is to compare the similarity of pixels between two images by moving the template window, and select the one with the largest similarity as the best correspondence. Traditional template-based methods use the intensity information, such as normalized cross-correlation (NCC) \cite{ma2010fully} and mutual information (MI) \cite{cole2003multiresolution}. NCC has difficulty dealing with nonlinear differences. MI describes the statistical information of the intensity distribution, which can cope with nonlinear differences to a certain extent. But it can't achieve the global optimum. Since the geometric structure and shape features are stable, Ye et al. proposed a dense local self-similarity shape descriptor based on them \cite{ye2017robust}. However, shape descriptors rely too much on the continuous shape and contour, and their generality is low. Later, Ye et al. proposed histogram of oriented phase congruency (HOPC) descriptor \cite{ye2017robust1} to extracts the robust phase congruency feature of images. But HOPC uses sparsely sampled grids, which is hard to accurately capture the local structural details. To this end, Ye et al. proposed the channel feature of directional gradient (CFOG) \cite{ye2019fast}. CFOG constructs descriptors at each pixel, which enhances the ability to describe the detailed structure, and improves the matching performance. Although template-based methods have been well developed, they need datasets to provide the initial matching locations with high accuracy. If there is a large deviation in the locations, their matching performance drops dramatically. Moreover, they are sensitive to large geometric transformations.
\subsection{Deep Learning-based Methods}		
With the development of deep learning, deep learning-based methods have been widely used in MIM. Zhao et al. proposed a Heterogeneous SuperPoint Network (HSPN) \cite{zhao2022heterogeneous}. They develop a strategy for training SuperPoint \cite{detone2018superpoint} detector that used in conjunction with heterogeneous homographic adaptation to generate pseudo-ground truth interest point labels for unlabeled images in a self-supervised fashion. Deng et al. proposed Recoupling Detection and Description for Multimodal Feature Learning (ReDFeat) \cite{deng2022redfeat}. They recouple independent constraints of detection and description of multimodal feature learning with a mutual weighting strategy, in which the detected probabilities of robust features are forced to peak and repeat, while features with high detection scores are emphasized during optimization. Wang et al. proposed a cross-modality perceptual style transfer network (CPSTN) that can convert visible images into pseudo infrared images to reduce the modal differences \cite{wang2022unsupervised}. CPSTN has good geometry preservation ability. Cao and Shi et al. proposed a Robust Deep Feature Matching method (RDFM) \cite{cao2023rdfm}. It extracts the deep features through a pre-trained VGG network, and designs a 4D convolution-based implementation of dense template matching, which can overcome the nonlinear radiation difference. Xu et al. pioneered a novel method to achieve multimodal image registration and fusion in a mutually reinforcing framework, termed as RFNet \cite{ye2018remote}. Registration is handled in a coarse-to-fine approach. The coarse registration corrects the global parallaxes between the images by TransNet and AffineNet, where TransNet is used to remove modal differences. Although deep learning-based methods show great potential, their universality and stability are poor due to the wide variety, complex scenes and insufficient training data of multimodal images.
\subsection{Feature-based Methods}	
The basic steps of feature-based methods are feature detection, description, matching and false match removal \cite{ma2022feature}. There are two kinds of feature-based methods for MIM:
\subsubsection{Spatial-Domain based Methods}	
The spatial-domain methods mainly use the gradient or intensity information in spatial-domain of images. Xiong et al. proposed a robust registration algorithm for optical and SAR images based on the adjacent self-similarity (ASS) \cite{xiong2022robust}. The ASS feature of the pixelwise feature representation is defined to quickly and finely capture the structural features of the image. They extract the minimum self-similarity map (SSM) and the index map for matching, which are robust to radiometric differences and speckles. Yao et al. proposed a method based on co-occurrence filtering (CoF) space matching (CoFSM) \cite{yao2022multi}, which constructs a new co-occurrence scale space based on CoF and extracts feature points in it. CoFSM uses HOG to construct descriptor, which can preserve edge information. However, its performance degrades in the presence of noises and distortions, or when the texture is discontinuous. In general, spatial-domain based methods give good results only for certain types of images.

\subsubsection{Frequency-Domain based Methods}
For different multimodal images, there are differences in NIDs and spatial information, so the universality of spatial-domain based methods is low. Compared with spatial-domain, frequency-domain information is more suitable for MIM. 2D log-Gabor filter (LGF) can extract frequency-domain information which is invariant to NIDs, so they are widely used, such as log-Gabor histogram descriptor (LGHD) \cite{aguilera2015lghd}, radiation-variation insensitive feature transform (RIFT) method \cite{li2019rift} and the histogram of orientation of weighted phase (HOWP) method \cite{zhang2023histogram}. RIFT first detects corner and edge feature points on phase congruency map via the maximum moment algorithm, and constructs maximum index descriptors based on LGF convolution image sequences. HOWP uses feature aggregation strategy to extract feature points, and establishes a new weighted phase orientation model to construct regularization-based log-polar descriptor, which can resist radiometric distortion and contrasting differences. The frequency-domain based methods are invariant to NIDs, but less robust to noises. Because the directional sensitivity and mutation of the phase extreme value affect the phase orientation features of descriptors, the frequency-domain based methods are very sensitive to the rotation angle and the matching performances of these methods decrease with the increase of image rotation angle.

In summary, the feature-based methods are robust to geometric deformations of images, and do not need the initial locations \cite{zhao2020image}. However, they rely on the repeatability of feature points. Therefore, when there are large nonlinear distortions and noise, the repeatability and localization accuracy of feature points will be reduced, and their matching performance will be significantly degraded.	

Based on the current research, the existing methods are still not able to fully solve three MIM challenges mentioned in introduction. Therefore, this paper proposes a robust method based on the frequency-domain information. In our work, FILER constructs the local energy response model to eliminate NIDs, and designs edge structure enhanced feature detector and convolutional feature weighted log-polar descriptor to overcome the problems of noises, local nonlinear geometric distortions and rotation transformation. Compared to state-of-the-art methods, FILER achieves significant improvement in performance.	
\section{Frequency-domain Information of Local Energy Response (FILER)}
We propose a novel and robust FILER algorithm for MIM task. In this section, we detail the principle of FILER.
\begin{figure*}[t]	
	\centering	
	\includegraphics[scale=0.23]{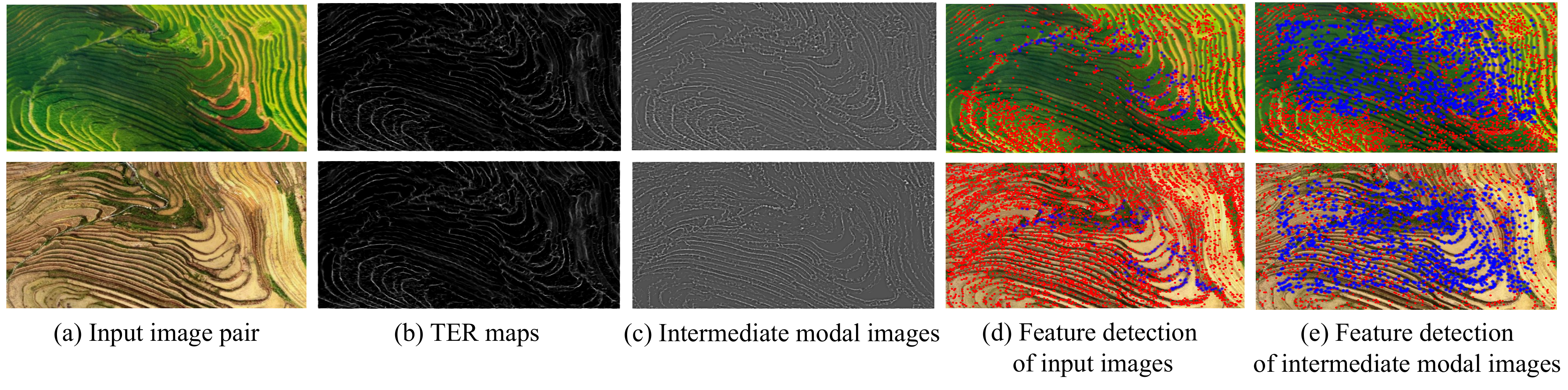}
	\caption{Feature detection. (a) is the input image pair from remote sensing cross-season image pairs. (b) and (c) are total energy response map and intermediate modal image $I_{out}$ of remote sensing cross-season image, respectively. (d) is the result of feature detection from the input images, and (e) is the result of feature detection from $I_{out}$, where the red points are the outliers without matching, and the blue '$*$' are the inliers successfully matched.}\label{fig1}
\end{figure*}
\subsection{Local Energy Response Model}
The intensity and gradient of multimodal images at same position are different, and feature detectors based on them are no longer applicable. In Section I, we analyze that multimodal images have complex NIDs due to multiple factors. To overcome NIDs, we hope to establish a model that can detect and describe the common features of them.

Extensive MIM studies have shown that the most common features between multimodal images are edges and bars. As shown in Figure~\ref{fig1}, we also find in our experiments that the correctly matched feature points are clearly concentrated in the edges and bars. Frequency-domain filter (FDF) can detect edges and bars by constructing local energy response, such as LGF, since step edges and bars have specific local phase properties which can be detected using filters in quadrature. Considering that the image is a 2D signal containing a wide frequency range, we construct the local energy response model by multi-scale and multi-orientation 2D FDF. First, we introduce the frequency-domain formula of the FDF:
\begin{equation}
	\begin{split}
		\label{equal1}
		FDF(f,\theta ,s,o) = &\exp \left( { - \frac{{\left( {\log \left( {f/{f_0}} \right)} \right)}}{{2\left( {\log \left( {{\sigma _f}/{f_0}} \right)} \right)}}} \right) \\
		& \cdot \exp \left( { - \frac{{\theta  - {\theta _{so}}}}{{2\sigma _\theta ^2}}} \right), 
	\end{split}
\end{equation}
where $ {f_0} $ is the center frequency of FDF, and $ {\theta _{so}} $ is the orientation angle. By adjusting $ {f_0} $, we can obtain multi-scale FDF. ${\sigma _f}$ and ${\sigma _\theta }$ are bandwidth of $f$ and the angular bandwidth of $\theta$, respectively. $s$ and $o$ are the wavelet scale and orientation of FDF used in the frequency-domain analysis, respectively.

The spatial-domain form of FDF has two components: even-symmetric filter and odd-symmetric filter \cite{kovesi2000phase}. Therefore, the spatial-domain formula of FDF is as follows:
\begin{equation}\label{equal2}
	FDF(x,y,s,o) = FD{F^{even}}(x,y,s,o) + j \cdot FD{F^{odd}}(x,y,s,o),
\end{equation}	
where $FD{F^{even}}(x,y,s,o)$ and $FD{F^{odd}}(x,y,s,o)$ are the real and imaginary parts of FDF at scale $s$ and orientation $o$, respectively. They also represent the even-symmetric and odd-symmetric wavelets, respectively.

Wavelets can detect discontinuous local features without intrinsic scale, since they can be reflected in the coefficient values of the wavelets. The wavelet response component $ {{E_{so}}\left( {x,y} \right)} $ and $ {{O_{so}}\left( {x,y} \right)} $ can be obtained by convolving the input image $ {{I_{in}}\left( {x,y} \right)} $ with the even-symmetric and odd-symmetric wavelets:
\begin{equation}\label{equal3}
	\left\{ {\begin{array}{*{20}{c}}
			{{E_{so}}\left( {x,y} \right) = {I_{in}}\left( {x,y} \right)*FD{F^{even}}(x,y,s,o)},\\
			{{O_{so}}\left( {x,y} \right) = {I_{in}}\left( {x,y} \right)*FD{F^{odd}}(x,y,s,o)}.
	\end{array}} \right. 
\end{equation}

The amplitude of response component in $o$ and $s$ is:
\begin{equation}\label{equal4}
	A{m_{s,o}}\left( {x,y} \right) = \sqrt {E_{so}^2\left( {x,y} \right) + O_{so}^2\left( {x,y} \right)}.  
\end{equation}

To make local energy response model make sense over a wide range of frequencies, we calculate the sum of ${E_{so}}\left( {x,y} \right) $ and $ {O_{so}}\left( {x,y} \right) $ of all scales in each orientation to form a relatively uniform spectral coverage of the signal. We can obtain the total response in each orientation: $\sum\limits_s {{E_{so}}\left( {x,y} \right)} $ and $\sum\limits_s {{O_{so}}\left( {x,y} \right)} $. Because the second-order nonlinearity of local energy form can detect many natural edges and contours with complex intensity distribution in images well, we define the local energy response in $o$ as:
\begin{equation}\label{equal5}
	Energ{y_o}\left( {x,y} \right) = \sqrt {{{\left( {\sum\limits_s {{E_{so}}\left( {x,y} \right)} } \right)}^2} + {{\left( {\sum\limits_s {{O_{so}}\left( {x,y} \right)} } \right)}^2}}. 
\end{equation}

At each point $(x, y)$ of the image, we can obtain a vector $Energy_{o\in [1, N_o]}$, one element of which represents one orientation of FDF. To detect the features in all orientations, the orientations of FDF are uniformly distributed in frequency plane. The multi-scale property of FDF can decompose the signal into frequency components of multi-scales, which can fully retain the details and overall structure information of the image, and is conducive to extract more effective features \cite{zhou2023attention, luo2023multiscale}. $Energy_o$ are the basis of features detector and descriptor. 
\subsection{Feature Detection} 
MIM is strongly affected by feature detector. If the accuracy of feature points are low, the matching results will not be good. We design a novel edge structure enhanced feature detector, which can overcome the noise interference and detect high quality feature points.
\subsubsection{Total Energy Response} 
We have established a local energy response model and obtained $Energy_o$. The total energy response $ET$ with large-scale edge structures is obtained by the following: (1) Normalize the local energy response to avoid the inconsistencies of their size standards. The specific operation is to divide $({\sum\nolimits_s {\sum\nolimits_o {A{m_{s,o}}\left( {x,y} \right) + \varepsilon } } })$, where ${\sum\nolimits_s {\sum\nolimits_o {A{m_{s,o}}\left( {x,y} \right)} } }$ is the sum of amplitudes in all orientations and scales and $\varepsilon$ is a small constant. (2) The noise compensation term $T_0$ is introduced to subtract the spurious response of the noise in $Energy_o(x, y)$. Note that noise compensation is performed independently in each orientation. (3) Add weighting function $W_o(x, y)$ to extend frequency distribution and enhance the localization of features, especially for smoothed features. The value of $W_o(x, y)$ is related to the frequency spread, and spurious responses can be reduced by lowering the weight where the frequency spread is narrow. The above process is expressed by the formula:	
\begin{equation}\label{equal6}
	ET\left( {x,y} \right) = \frac{{\sum\nolimits_o {{W_o}\left( {x,y} \right)\left| {Energ{y_o}\left( {x,y} \right) - {T_0}} \right|} }}{{\sum\nolimits_s {\sum\nolimits_o {A{m_{s,o}}\left( {x,y} \right) + \varepsilon } } }},
\end{equation}	
where $\varepsilon$ is a small constant to prevent the denominator from being 0. $|\cdot|$ guarantees that the output is non-negative.

\subsubsection{Edge Structures Enhancement} 
$ET$ eliminates redundant and useless smaller scale textures, but retains many obvious large-scale edge structures. Because compared with large-scale structures, the repeatability of small-scale texture in multimodal images is very low, which is not conducive to the accuracy of feature detection. However, due to the interference of noises and outliers, such as multiplicative and speckle noise in remote sensing images, part of the large-scale edge structures are blurred and important details are destroyed.

In order to enhance the blurred edge structures, we design an edge structure enhanced detector to automatically refine the edges that need to be retained. We use $ ET\left( {x,y} \right)$ as the guidance image because $ ET\left( {x,y} \right)$ can identify edge structures. The specific steps are:
\begin{equation}
	\begin{split}
		\label{equal7}
		&J(j) = \frac{1}{{{W_j}}}\sum\limits_{i \in N(j)} {{G_d}(i - j){G_r}(i - j)} {I_{in}}(i), \\
		&{W_j} = \sum\limits_{i \in N(j)} {\exp \left( { - \frac{{{{\left\| {j - i} \right\|}^2}}}{{2\sigma _s^2}} - \frac{{{{\left\| {ET(j) - ET(i)} \right\|}^2}}}{{2\sigma _r^2}}} \right)},	
	\end{split}
\end{equation}
where ${I_{in}}(i)$ and $J(j)$ denote the input and output images, respectively. $j$ and $i$ denote index pixel coordinates. $N(j)$ is the set of neighboring pixels of $j$ in spatial-domain of the image, and $W_j$ denotes normalization operation.

The spatial distance function $G_d$ in Eq.~(\ref{equal7}) sets the weights according to the distance between pixels, while the structural guidance function $G_r$ sets the weights according to the $ET(x, y)$. They are defined as:	
\begin{equation}
	\begin{split}
		\label{equal8}
		&{G_d}(i - j) = \exp \left( { - \frac{{{{\left\| {j - i} \right\|}^2}}}{{2\sigma _s^2}}} \right), \\
		&{G_r}(i - j) = \exp \left( { - \frac{{{{\left\| {ET(j) - ET(i)} \right\|}^2}}}{{2\sigma _r^2}}} \right),
	\end{split}
\end{equation}
where ${\sigma_s}$ and ${\sigma_r}$ control the spatial weight and structural weight respectively. 

Eq.~\eqref{equal7} can be understood as filtering $I_{in}$ guided by $ ET\left( {x,y} \right)$. Through the action of structural guidance function, the filter performs less averaging around the edge structures, making them sharper. At this time, the preservation of edges depends not on their scale, but on the distribution of total energy. However, it inevitably smooths some detail structures.

To restore the detail structures mistakenly smoothed in the filtering process, we take the difference between $J(j)$ and $ ET\left( {x,y} \right)$ as the final image structure map: 
\begin{equation}
	\label{equal9}
	{I_{out}} = J(j) - ET(j).
\end{equation}

We use $I_{out}$ as the intermediate modal image which has more common information to improve FAST \cite{rosten2008faster} for detect. Figure~\ref{fig1} is an example of feature detection. (c) retains obvious large-scale edge and detail structures, and also reduces the influence of noise. By comparing (d) and (e), we can find that gradient-based feature detector cannot cope with NIDs, and its feature points have low repeatability. The feature points detected based on $I_{out}$ are robust to NIDs, which reflect the common features between two images well.	
\subsection{Feature Matching}
\subsubsection{Feature Description}
\begin{figure}[h]	
	\centering	
	\includegraphics[scale=0.23]{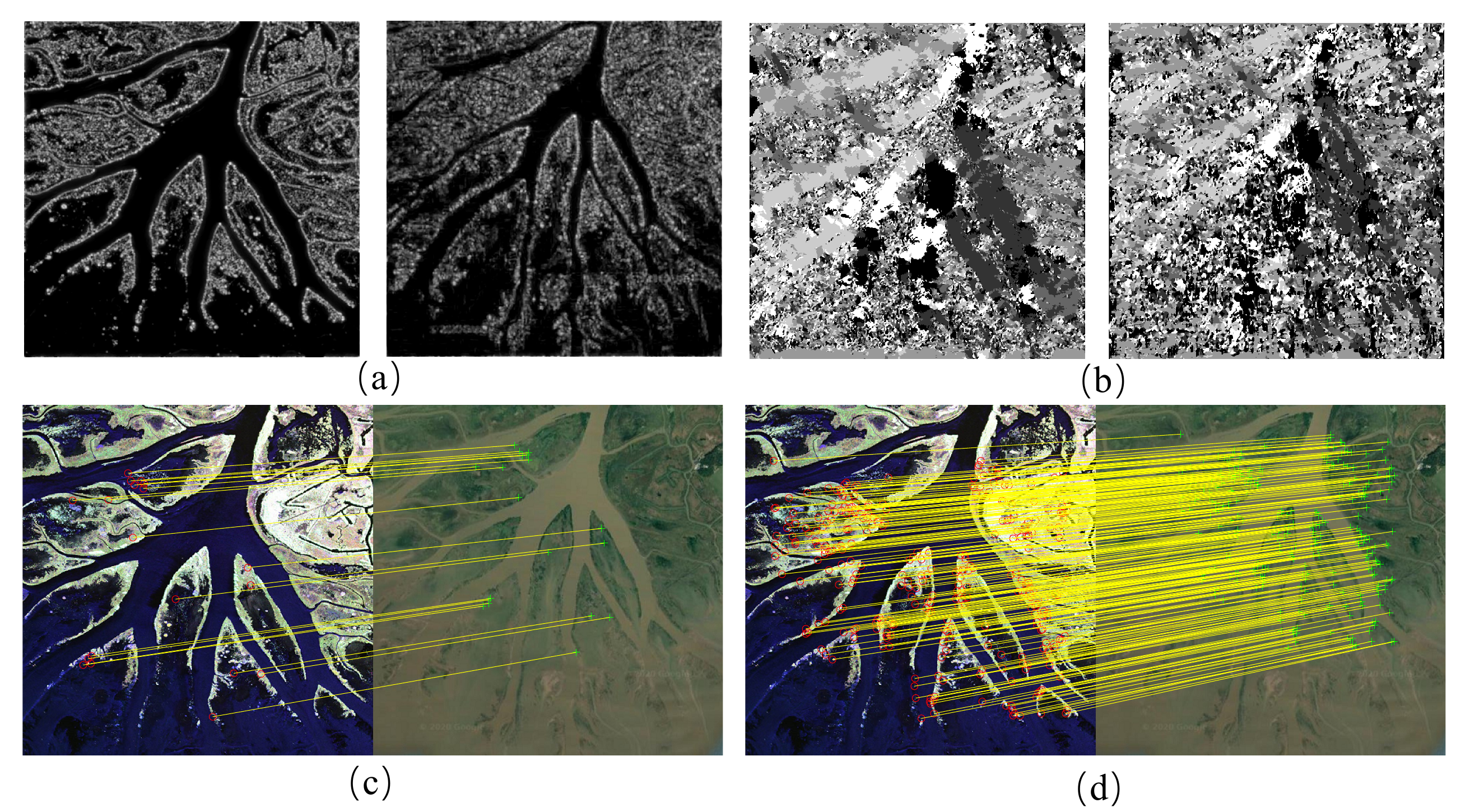}
	\caption{The matching results of descriptors constructed based on total energy response and energy manifold vector field. The images in (a) are the total energy response maps of the SAR-optical image pair and the images in (b) are the energy manifold vector fields of them. (c) is the matches established by using (a) as the feature description map, and (d) is the matches established by using (b).}\label{fig2}
\end{figure}
After feature detecting, we need to describe them to distinguish and match. $ Energ{y_o}( {x,y} ) $ contains lots of common features and is well suited for feature description. From the previous analysis, we know that $ET$ has strong robustness to NIDs, but the feature description based on $ET$ does not have good matching results in Figure~\ref{fig2} (c). This is because most pixel values of $ET$ are close, and many details are lost after filtering. These details may interfere with feature detection, but are useful for feature description. Therefore, $ET$ is not conducive to feature description. We construct the energy manifold vector field of the image by taking the maximum orientation of $Energ{y_{o\in [1,N_o]}}$, which contains a wealth of texture details, and we build the descriptor based on it. However, local nonlinear geometric distortions destroy the texture details of energy manifold vector field, and lead to the increase of false matches. So we design a convolutional feature weighting method to improve the robustness of descriptor.

If the $Energy_o(x,y)$ in each orientation is a local energy response layer, the layers in all orientations can form a sequence. In other words, by ordering these layers in the dimension of $o$, a 3D cube is obtained. Subsequently, we use convolutional feature weighting in it. The 3D convolution kernel consists of a 2D Gaussian kernel in the $x$ and $y$ dimensions and a ${d_o} = {[1,8,1]^T}$ kernel in $o$ dimension. The processing of convolutional feature weighting is:
\begin{equation}
	\begin{split}
		\label{equal10}
		&Energy_o^{\sigma}(x,y)=G_{xy}^\sigma*Energy_o(x,y), \\
		&T_o(x,y)=d_o*Energy_o^{\sigma}(x,y),
	\end{split}
\end{equation}
where $T_o$ is the result of convolutional feature weighting, and $ G_{xy}^\sigma $ is the Gaussian convolution kernel, which suppresses noise by reducing the contribution of energy that far from the region center. $d_o$ is the convolution kernel in dimension $o$. It smooths the gradients and mitigates the impact of orientation deformation caused by local nonlinear geometric distortions.
\begin{figure}[t]	
	\centering	
	\includegraphics[scale=0.18]{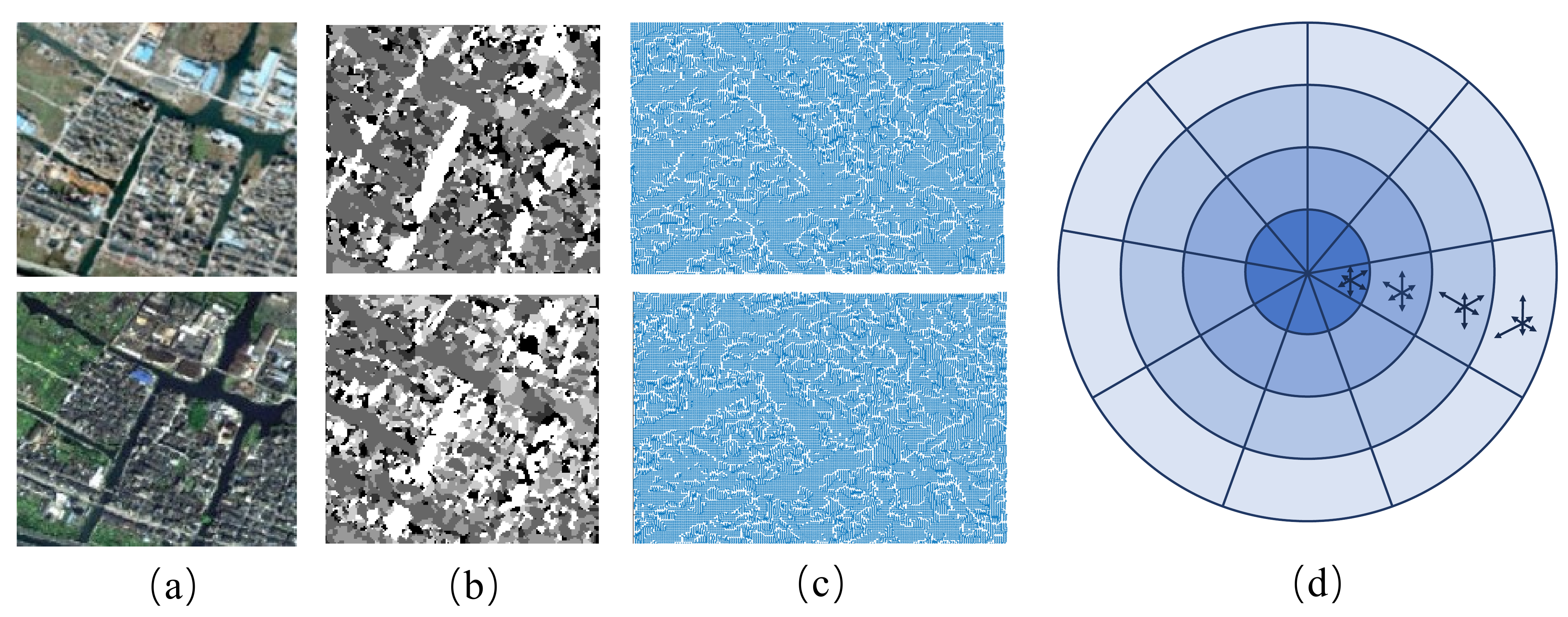}
	\caption{The energy manifold vector field diagram and log-polar grid of feature description map. (a) is the remote sensing optical cross-temporal image pair. (b) is the feature description map of (a) and (c) is the manifold energy vector field of (b). (d) is a circular log-polar grid with 36 location bins and each location bin contains a distribution histogram with $N_o$ bins.}\label{fig3}
\end{figure}

Next, we normalize ${T_o}\left( {x,y} \right)$ in dimension $o$:
\begin{equation}
	\label{equal11}
	{T_o}\left( {x,y} \right) = \frac{{{T_o}\left( {x,y} \right)}}{{\sqrt {\sum\nolimits_{o = 1}^{{N_o}} {{{\left| {{T_o}\left( {x,y} \right)} \right|}^2}}  + \varepsilon } }},
\end{equation}
where $\varepsilon $ is a constant and ${N_o}$ is the number of orientations.

After the weighting of convolutional feature, we can obtain a multi-channel information map $\{ {{T_o}( {x,y} )} \}_{o = 1}^{N_o}$ by ordering ${{T_o}( {x,y} )}$. Next, we use maximum orientation method to assign dominant orientation from $\{ {{T_o}( {x,y} )} \}_{o = 1}^{N_o}$ and construct the energy manifold vector field. The basic operation is to find $o_{\max }$ that corresponds to the maximum value of $\{ {T_o}(x,y) \}_{o = 1}^{N_o}$ i.e., $o_{\max } = \mathop {\arg \max }\limits_o \{ {\{ {{T_o}( x_i,y_i )} \}_{o = 1}^{N_o}} \}$. Finally, we use ${o_{\max }}$ of all pixels as feature description map. On the physical level, $o_{\max }$ is the most significant orientation corresponding to the local energy of each pixel. As shown in Figure~\ref{fig3} (c), the feature description map is actually the energy manifold vector field formed by $o_{\max }$. Our feature descriptor can establish a greater number of matches that are more widely distributed from the comparison in Figure~\ref{fig2}.

Now, we construct the log-polar structure descriptor based on $o_{\max }$. Because it is more robust than the square patch descriptor when the images have local nonlinear geometric distortions \cite{shechtman2007matching}, \cite{bellavia2017rethinking}. First, a circular patch centered around the feature points is taken and divided into a log-polar grid of 4 radial bins and 9 angular bins, resulting in a total of 36 location bins. Figure~\ref{fig3} (d) is a log-polar grid with 36 location bins. We construct a distribution histogram with $N_o$ bins for each location bin and obtain the final descriptor of size $4 \times 9 \times {N_o}$ that concatenates the distribution histograms of all the location bins. Finally, we normalize the feature descriptor vector to achieve the invariance of brightness.
\subsubsection{Establishment of Putative Matches and False Match Removal} 
We use the nearest neighbor (NN) matching technology and the similarity measure of Euclidean distance to establish putative matches which are one-to-one. The fast sample consistency (FSC) algorithm \cite{wu2014novel} is used to remove false matches with fewer iterations. If the match $(X,Y)$ meets the following equation, it will be judged as an inlier:
\begin{equation}
	\label{equal110}
	Y-H*X<\delta ,
\end{equation}
where $H$ is the affine transformation matrix estimated by FSC. $\delta$ is the threshold to select the true matches with small deviations.
\begin{figure}[t]	
	\centering	
	\includegraphics[scale=0.25]{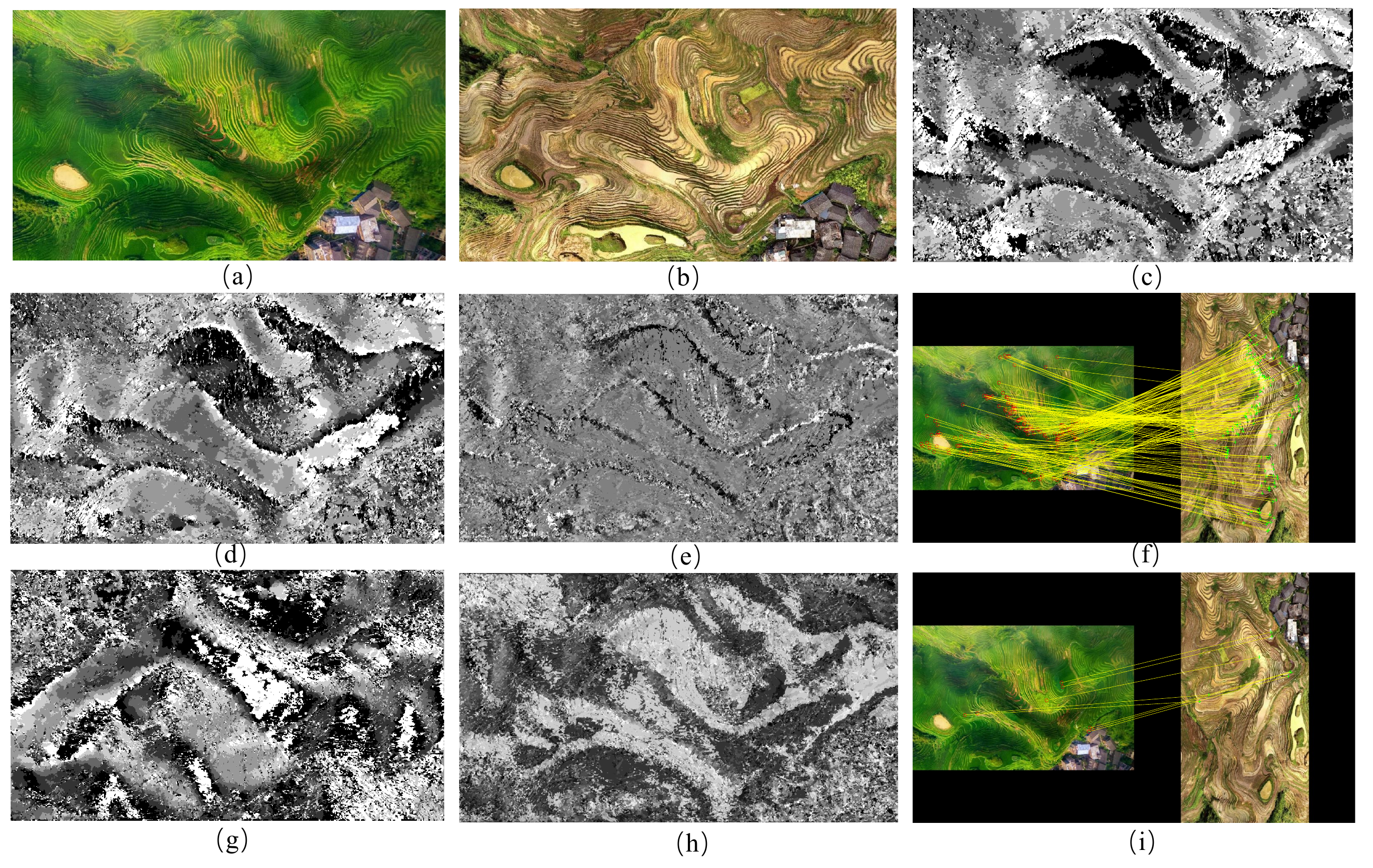}
	\caption{Sensitivity of feature description map to rotation angle. (a) and (b) are remote sensing cross-season image pair. (c) and (d) are the feature description maps of (a) and (b), respectively. (g) is the feature description map of (b) rotated $90^{\circ}$ counterclockwise. (e) is the difference map between (c) and (d). (h) is the difference map between (c) and (g). (f) is the matching result based on (c) and (d). (i) is the matching result based on (c) and (g).}\label{fig4}
\end{figure}
\subsection{Implementation of Rotation Invariance}
In a large number of MIM experiments, we find the rotation sensitivity problem of frequency-domain based methods. Because the directional sensitivity and mutation of the phase extreme value affect the phase orientation features of descriptors, the frequency-domain based methods are very sensitive to the rotation angle. If there is a large rotation angle in multimodal images, their matching results will be poor, as shown in Figure~\ref{fig4} (i). This is because the feature description maps of frequency-domain based methods are closely related to the angular component of FDF (the second term of Eq.~(\ref{equal1})). When there is a rotation in two images, the start angles of their angular components are different, which leads to differences in feature description maps. Matching based on different feature description maps is usually unsuccessful. From Figure~\ref{fig4} difference map (h), we see that (c) and (g) are different, because there is a $90^{\circ}$ difference in $\theta_{so}$. The result (i) is failed. In contrast, the difference map (e) between (c) and (d) is small, because there is a $0^{\circ}$ difference in $\theta_{so}$. The result (f) is excellent. In order to achieve rotation invariance, we design main direction method to reduce the difference in feature description maps due to rotation.

In feature detection, we get ${I_{out}}$. We can use the ${I_{out}}$ to extract gradient direction of each pixel, and rotate the patches of descriptors according to the main direction.

Firstly, the multi-dimensional filter operator $H = [-1,0,1]$ is used to filter ${I_{out}}$, and the vertical and horizontal gradient maps of the image are calculated:
\begin{equation}	
	\label{equal12}
	\begin{split}
		{G_x}\left( {x,y} \right) = P\left( {x + 1,y} \right) - P\left( {x - 1,y} \right),\\
		{G_y}\left( {x,y} \right) = P\left( {x,y + 1} \right) - P\left( {x,y - 1} \right),
	\end{split}
\end{equation}
where ${G_x}\left( {x,y} \right)$, ${G_y}\left( {x,y} \right)$ and $P\left( {x,y} \right)$ represent the horizontal, vertical gradient and pixel value of the feature description map at $ \left( {x,y} \right) $, respectively. The pixel gradient amplitude $G(x,y)$ and gradient direction $Angle(x,y)$ at $ \left( {x,y} \right) $ are:
\begin{equation}	
	\label{equal13}
	\begin{split}
		G\left( {x,y} \right) = \sqrt {{G_x}{{\left( {x,y} \right)}^2} + {G_y}{{\left( {x,y} \right)}^2}}, \\
		Angle\left( {x,y} \right) = \arctan \left( {\frac{{{G_y}\left( {x,y} \right)}}{{{G_x}\left( {x,y} \right)}}} \right).
	\end{split}
\end{equation}
\begin{figure*}[t]	
	\centering	
	\includegraphics[scale=0.22]{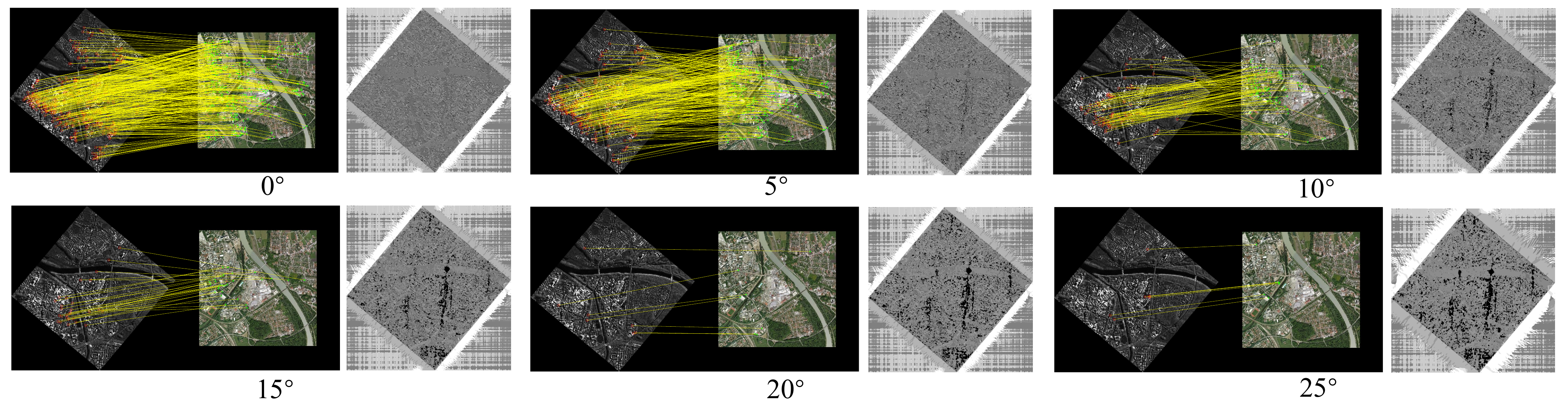}
	\caption{Matching results and difference maps at different angle errors $[0^{\circ}, 5^{\circ}, 10^{\circ}, 15^{\circ}, 20^{\circ}, 25^{\circ}]$. The left of each group of images is the matching result, and the right is the difference map of feature description maps of $50^{\circ}$ rotation angle.}\label{fig5}
\end{figure*}

Next, we draw a circular patch of radius $r$ with the feature point as the center of circle, and divide the $n$ angles equally. Subsequently, the sum of gradient amplitudes in each angle interval is calculated. The angle is used as abscissa and the sum of gradient amplitudes in each angle interval is used as ordinate to compute the histogram. The direction with the largest ordinate (or with the extreme value) is selected as the main direction of feature point. Then we rotate the patch of feature descriptor according to the main direction.

We find that the results of rotated images using the main direction method are rough, with most corresponding points having the error of more than 3 pixels. Therefore, it is not feasible to use them as the final results. However, we can adjust the parameters to obtain a small number of matching points with high accuracy, and use them to estimate the total rotation angle $\theta$. $\theta$ can be used to rotate the descriptor's patch and eliminate the differences of $\theta_{so}$ in the rotated images i.e., $\theta_{so}^1-\theta=\theta_{so}^2$ ($\theta_{so}^1$ and $\theta_{so}^2$ are $\theta_{so}$ of the two images, respectively).
\begin{table}[t]
	\small
	\renewcommand\arraystretch{0.8}   
	\centering
	\caption{The NCM and CMR under different angle errors.}
	\label{table1}
	\setlength{\tabcolsep}{5mm}{  
		\begin{tabular}{ccccccc}
			\toprule
			Angle error ($^\circ$)& 0& 5& 10& 15& 20& 25 \\
			\midrule
			NCM& 367& 268& 103& 22.0& 1.00&0\\
			CMR($\%$)&100&99.6& 100&95.7&14.0&0\\
			\bottomrule	
		\end{tabular}
	}
\end{table}

The error of estimated rotation angle is within $10 ^{\circ}$. We test in representative multimodal images to explore what range of the angle error can satisfy the matching accuracy and number requirements. As shown in Figure~\ref{fig5} and Table~\ref{table1}, we set different angle errors and match them, respectively. As the error increases, the matching performance decreases, but when the error is within $10 ^{\circ}$, the result is acceptable. When the error is greater than $10 ^{\circ}$, the number of correct matches (NCM) and correct match ratio (CMR) decrease significantly, and the difference map becomes obvious.

\section{Experimental Results}
To test FILER, we conduct qualitative and quantitative experiments on three multimodal datasets and compare it with 7 state-of-the-art algorithms: CoFSM \cite{yao2022multi}, HOWP \cite{zhang2023histogram}, HSPN \cite{zhao2022heterogeneous}, ReDFeat \cite{deng2022redfeat}, RDFM \cite{cao2023rdfm}, RIFT \cite{li2019rift}, and CPSTN+SIFT \cite{wang2022unsupervised}. To be fair, all codes are taken from the download sites provided by authors. Each algorithm has its own feature detector, so we test their feature detectors. In feature matching, all algorithms use their own feature detectors and descriptors, and the same false match removal method. CPSTN is only trained on datasets with infrared and visible images. Therefore, our method is compared with it on infrared image (IR)-optical and visible-IR image pairs. We pair CPSTN with SIFT to test. Because the number of multimodal image pairs used for experiments is small and does not meet the training requirements of CPSTN, we use the pretrained models provided by the official implementation for testing. We design ablation experiments to verify the effect of each module of FILER. Furthermore, we also test the rotation invariance of FILER and its performance in image registration and fusion. For the best performance, we finetune the parameters of each method. 
\subsection{Datasets}
\begin{figure*}[t]	
	\centering	
	\includegraphics[scale=0.19]{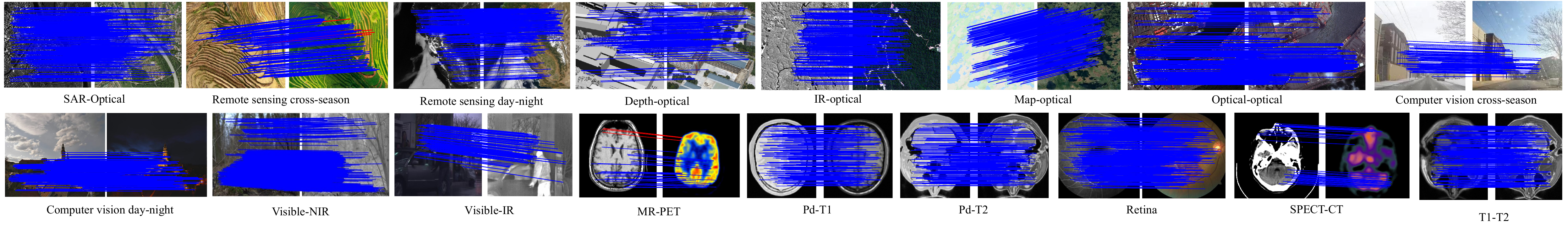}
	\caption{Typical images of all categories and their matching results (The blue lines are true matches, and the red lines are fasle matches).}\label{fig7}
\end{figure*}

The datasets contain three categories, which are remote sensing, medical and computer vision \cite{jiang2021review} dataset. There are 164 image pairs, and the typical images are shown in Figure~\ref{fig7}.

(1) The remote sensing dataset has 7 types of image pairs, which are unmanned aerial vehicle (UAV) cross-season, optical day-night, light detection and ranging (LiDAR) depth-optical, IR-optical, map-optical, optical cross-temporal, and SAR-optical image pairs \cite{yao2022multi}, respectively. These multimodal images have different imaging devices, imaging times and types, and are subject to noises and local nonlinear geometric distortions \cite{fan2012registration}. 

(2) The computer vision dataset has 4 types of image pairs, which are visible-IR, visible-near-infrared (NIR) \cite{brown2011multi}, visible cross-season \cite{ma2021image}, and day-night image pairs \cite{mishkin2015wxbs}, respectively. These images suffer from several problems such as large geometric deformations, low quality, noise, occlusion, abnormal illumination and low texture.

(3) The medical dataset has 6 types of image pairs, which are magnetic resonance imaging (MRI)-positron emission tomography (PET), single photon emission computed tomography (SPECT)-computed tomography (CT), MRI T1-T2, proton density (PD)-T2, PD-T1 \cite{cocosco1997brainweb}, and retinal image pairs \cite{ma2019locality}, respectively. Medical images of different modes contain significantly different information (structural or functional information) \cite{maes2003medical}.

All image pairs are manually marked with 15 to 20 matching landmarks, which is used to evaluate the accuracy of registration. The authors also provide the affine transformation matrix $H$ as groundtruth. We calculate the residuals between the matching points and groundtruth, and take the matches with the residuals less than 3 pixels as true matches.

\subsection{Parameter Analysis and Setting}
There are three key parameters of FILER, which are $N_o$, $N_s$ and $r$. $N_o$ and $N_s$ are the number of orientations and scales of 2D FDF, respectively. If $N_s$ and $N_o$ are large, the feature description map contains more details, and the performance of FILER will be improved to a certain extent, but the performance improvement is limited and the run time will increase. FILER constructs descriptors through circular patches, where $r$ denotes the radius of the circular patches. The radius $r$ determines how much information is contained in descriptor. If it is too small, the information will be insufficient. If it is too large, it will inevitably contain error information due to local nonlinear geometric distortions. To get the best results, we need to choose the best $N_s$, $N_o$, and $r$. We design three experiments for three parameters. The experiments use a mixed dataset consisting of 90 multimodal image pairs, where the images are randomly selected from the datasets in Section IV-A. NCM, CMR and runtime are the evaluation indexes.

Table~\ref{table2}, Table~\ref{table3} and Table~\ref{table4} show the results. In Table~\ref{table2}, as $N_o$ increases, NCM and CMR improve, but the runtime also increases. When $N_o>6$, the increase of NCM and CMR slow down or even decrease, so to balance the performance and efficiency of FILER, we set $N_o=6$. In Table~\ref{table3}, FILER has the best performance with $N_s=4$, so we set $N_s=4$. In Table~\ref{table4}, NCM decreases very quickly with $r$ when $r>20$, and CMR decreases with $r$ when $r>50$. Considering all 3 evaluation indexes, $r = 40$ is the best. Now, we have the best parameter settings: $ N_s =4, N_o =6, r=40$. The parameter settings are kept the same in all subsequent experiments.
\begin{table}[t]
	\small
	\renewcommand\arraystretch{0.8}   
	\centering
	\caption{The result of different parameter $N_o$ (fix $N_s=4$, $r=40$).}
	\label{table2}
	\setlength{\tabcolsep}{5mm}{  
		\begin{tabular}{ccccccccccc}
			\toprule
			Index& $N_o=4$& $N_o=5$& $N_o=6$& $N_o=7$& $N_o=8$ \\
			\midrule
			NCM& 156.73& 169.13& 174.31& \textcolor{red}{\textbf{174.43}}& 169.43\\
			CMR($\%$)&88.139&88.246& \textcolor{red}{\textbf{90.410}}&89.460&87.762\\
			Runtime(s)& 9.1288&10.042 &10.527&11.086&11.330\\		
			\bottomrule	
		\end{tabular}
	}
\end{table}
\begin{table}[t]
	\small
	\renewcommand\arraystretch{0.8}   
	\centering
	\caption{The result of different parameter $s$ (fix $N_o=6$, $r=40$).}
	\label{table3}
	\setlength{\tabcolsep}{5mm}{  
		\begin{tabular}{ccccccccccc}
			\toprule
			Index&$N_s=2$& $N_s=3$& $N_s=4$& $N_s=5$& $N_s=6$ \\
			\midrule
			NCM& 152.57& 168.27& \textcolor{red}{\textbf{174.31}}& 162.47& 153.33\\
			CMR($\%$)&77.957&88.812& \textcolor{red}{\textbf{90.410}}&87.272&88.049\\
			Runtime(s)& 9.5232&9.8043 &10.527&10.918&11.197\\		
			\bottomrule	
		\end{tabular}
	}
\end{table}
\begin{table}[t!]
	\small
	\renewcommand\arraystretch{0.8}   
	\centering
	\caption{The result of different parameter $r$ (fix $N_s=4$, $N_o=6$).}
	\label{table4}
	\setlength{\tabcolsep}{3.5mm}{  
		\begin{tabular}{ccccccccccc}
			\toprule
			Index& r=10& r=20& r=30& r=40& r=50& r=60 \\
			\midrule
			NCM&139.93& \textcolor{red}{\textbf{203.97}}& 196.43& 174.31& 145.77& 102.64\\
			CMR($\%$)&51.362&81.938& 84.703&90.410&\textcolor{red}{\textbf{94.313}}& 86.351\\
			Runtime(s)& 9.7567&10.527 &10.560&10.786&12.089&12.922\\		
			\bottomrule	
		\end{tabular}
	}
\end{table}
\subsection{Performance Testing of Feature Detector}
We compare FILER with FAST, and the feature detectors of CoFSM, HOWP, HSPN, ReDFeat, RDFM, RIFT and CPSTN+SIFT algorithms. The evaluation indexes for detector are number of matches $N_c$, repeatability $Rep$, stability ratio of matching number $ Ratio({N_c}) $ and stability ratio of repeatability $ Ratio(Rep) $ \cite{mikolajczyk2001indexing}. $N_c$ and $Rep$ are defined as:
\begin{equation}	
	\label{equal17}
	\begin{split}
		&{N_c} = \left| {\left\{ {\left\| {{x_i} - H{y_i}} \right\| < 3} \right\}_{i = 1}^{{N_0}}} \right|, \\
		&Rep = \frac{{{N_c}}}{{({N_1} + {N_2})/2}},
	\end{split}
\end{equation}
\begin{table}[t]
	\small
	\renewcommand\arraystretch{0.7}   
	\centering
	\caption{The test results of the feature detector.}
	\label{table5}
	\setlength{\tabcolsep}{5.2mm}{  
		\begin{tabular}{ccccccccccc}
			\toprule
			Method &$N_c$ & $Rep(\%)$& $Ratio(N_c)$& $Ratio(Rep)$ \\
			\midrule
			FAST \cite{rosten2008faster}& 129.6& 2.664 &30.00 &3.333\\
			CoFSM \cite{yao2022multi}& 205.2& 15.90& 61.44& 54.55\\
			HOWP \cite{zhang2023histogram}& 540.9& 18.36& 94.44& 66.67\\
			HSPN \cite{zhao2022heterogeneous}& 216.4& 18.40&48.89& 62.22\\
			ReDFeat \cite{deng2022redfeat}& 502.7& 13.24& 76.92& 38.46\\
			RDFM \cite{cao2023rdfm}& 447.4&17.17& 88.89& 62.22\\				
			RIFT \cite{li2019rift}& 470.6 &17.07& 83.33& 61.11\\
			FILER& \textcolor{red}{\textbf{544.5}}& \textcolor{red}{\textbf{19.92}}& \textcolor{red}{\textbf{100.0}}& \textcolor{red}{\textbf{72.22}}\\
			\bottomrule	
		\end{tabular}
	}
\end{table}
\begin{table}[h]
	\small
	\renewcommand\arraystretch{0.7}   
	\centering
	\caption{The test results of the feature detector of FILER and CPSTN+SIFT in visible-IR and IR-optical image pairs.}
	\label{table6}
	\setlength{\tabcolsep}{4mm}{  
		\begin{tabular}{ccccccccccc}
			\toprule
			Method &$N_c$ & $Rep(\%)$& $Ratio(N_c)$& $Ratio(Rep)$ \\
			\midrule
			CPSTN+SIFT \cite{wang2022unsupervised}& 6.933& 0.310 &6.667 &6.667\\
			FILER& \textcolor{red}{\textbf{90.06}}& \textcolor{red}{\textbf{2.173}}& \textcolor{red}{\textbf{26.67}}& \textcolor{red}{\textbf{12.67}}\\
			\bottomrule	
		\end{tabular}
	}
\end{table}
where ${x_i}$ and ${y_i}$ are the homogeneous coordinates of the $i^{th}$ feature points of $I_1$ and $I_2$, respectively; $H$ is the affine transformation matrix (groundtruth); ${N_1}$ and ${N_2}$ are the number of feature points in $I_1$ and $I_2$, respectively; ${N_0}$ is the number of feature points that can be matched. $\left| {\left\{ {\left\| {{x_i} - H{y_i}} \right\| < 3} \right\}_{i = 1}^{{N_0}}} \right|$ represents the number of matches satisfying $\left\{ {\left\| {{x_i} - H{y_i}} \right\| < 3} \right\}_{i = 1}^{{N_0}}$ in the putative matches.

$Ratio({N_c})$ and $Ratio(Rep)$ are used to measure the stability of feature detector:
\begin{equation}	
	\label{equal18}	
	\begin{array}{*{20}{c}}
		{Ratio({N_c} > 100) = \frac{{\left| {\left\{ {N_c^i > 100} \right\}_{i = 1}^{{N_{im}}}} \right|}}{{{N_{im}}}}},\\
		{Ratio(Rep > 10\% ) = \frac{{\left| {\left\{ {Re{p^i} > 10\% } \right\}_{i = 1}^{{N_{im}}}} \right|}}{{{N_{im}}}}},
	\end{array}	
\end{equation}
where ${N_{im}}$ is the total number of image pairs in tested dataset. $Ratio({N_c} > 100)$ represents the proportion of images with $N_c$ greater than 100; $Ratio(Rep > 10\% )$ represents the proportion of images with repeatability $Rep$ greater than 10$\%$.

There are NIDs in multimodal images, so many feature points are easily eliminated for they are not significant enough to construct effective descriptors. As a result, the repeatability of feature detector is low, which indicates that extracting effective feature points is a challenging task. We still use the mixed dataset in Section IV-B for testing, and the results are shown in Table~\ref{table5}. Both FILER and HOWP have good comprehensive performance, but the $N_c$, $Rep$, $Ratio(N_c)$ and $Ratio(Rep)$ of FILER are higher, which indicates that the feature detector of FILER is more stable. The overall performance would be poor if we directly use the FAST to detect feature points in the original images. Table~\ref{table6} shows the feature detection results of our FILER and CPSTN+SIFT in visible-IR and remote sensing IR-optical image pairs, which indicates that the FILER has great advantages. The edge structure-enhanced feature detector of FILER transforms the original images into the intermediate modal images, which not only effectively removes noise, but also preserves sufficient common information. Therefore, it has the best overall performance.
\begin{table}[t]
	\fontsize{10pt}{10pt}\selectfont
	\renewcommand\arraystretch{0.8}   
	\centering
	\caption{The result of ablation experiment.}
	\label{table6-1}
	\setlength{\tabcolsep}{5mm}{  
		\begin{tabular}{ccccccccccc}
			\toprule
			Method &NCM & CMR($\%$) & Runtime(s)\\
			\midrule
			Phase Congruency Model& 115.5& 76.78 &15.49\\
			Local Energy Response Model& 140.0& 78.31& 7.169\\		
			\makecell[c]{Local Energy Response Model +\\log-polar descriptor}& 198& 80.30& 11.67\\
			\makecell[c]{Local Energy Response Model +\\edge structure enhancement} & 216.0& 81.26& 27.80\\
			\makecell[c]{Local Energy Response Model +\\convolutional feature}& 202.7& 81.48& 13.46\\
			\makecell[c]{Local Energy Response Model +\\All}& \textcolor{red}{\textbf{254.5}}& \textcolor{red}{\textbf{84.88}}&18.70\\

			\bottomrule	
		\end{tabular}
	}
\end{table}

\subsection{Ablation Experiment}
\begin{figure}[t]	
	\centering	
	\includegraphics[scale=0.23]{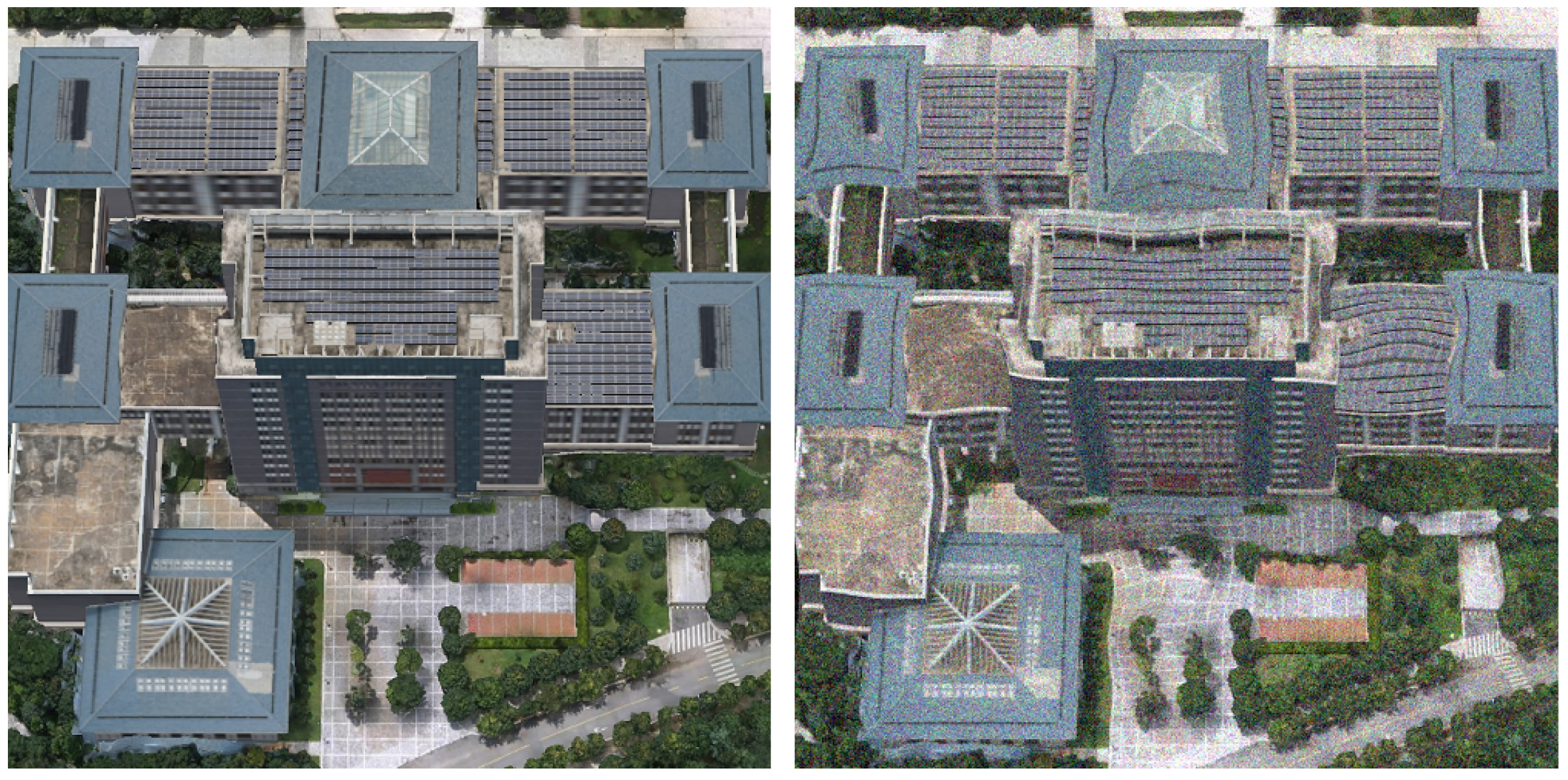}
	\caption{The image with noise and local nonlinear geometric distortions. The left is the original image, and the right is the image with noise and local nonlinear geometric distortions.}\label{fig10}
\end{figure}

In order to demonstrate the advantages of local energy response model, feature detector and descriptor in FILER, we design an ablation experiment. First, we randomly select 30 pairs of multimodal images for the experiment and add Gaussian noise (mean: 0, variance: 0.05) and local nonlinear geometric distortions to them, as shown in Figure~\ref{fig10}. We set up 6 comparison groups: phase congruency model \cite{li2019rift}, local energy response model, local energy response model + log-polar descriptor, local energy response model + edge structure enhancement, local energy response model + convolutional feature and local energy response model + all (FILER). The results of the ablation experiment in Table~\ref{table6-1} show that: (1) the local energy response model can effectively improve the feature matching performance; (2) the edge structure enhanced feature detector and convolutional feature weighted log-polar descriptor can enhance the robustness against noises and local nonlinear geometric distortions, and improve the matching accuracy. This indicates that the advantage of FILER lies in the design of local energy response model, feature detector and feature descriptor.
\subsection{Robustness Test with Different Noises}
\begin{table}[h]
	\small
	\renewcommand\arraystretch{0.8}   
	\centering
	\caption{Robustness test with different noises.}
	\label{table6-3}
	\setlength{\tabcolsep}{5mm}{  
		\begin{tabular}{ccccccccccc}
			\toprule
			Noise &NCM & CMR($\%$) & Runtime(s)\\
			\midrule
			Gaussian noise& 487.0& 84.79 &16.79\\
			Multiplicative noise& 444.7& 88.17& 29.69\\
			Poisson noise & 529.3& 88.27& 18.84\\
			Salt-and-pepper noise & 463.3& 85.49& 15.98\\
			Original images (w/o noise)& 602.7& 88.23& 17.00\\
			\bottomrule	
		\end{tabular}
	}
\end{table}	
To test the robustness of FILER with different noises, we randomly select 30 pairs of multimodal images. We apply Gaussian noise (mean: 0, variance: 0.01), multiplicative noise (variance: 0.01), Poisson noise, and salt-and-pepper noise (noise density: 0.04) to these images. As shown in Table~\ref{table6-3}, the NCM and CMR of images with different noise have decreased compared with the original image, but to a small extent, and their $(NCM, CMR)$ is greater than $(400, 80)$. The results show that FILER has good robustness to different noises because of the local energy response model and edge structure enhanced detector.

\begin{figure*}[t]	
	\centering	
	\includegraphics[scale=0.32]{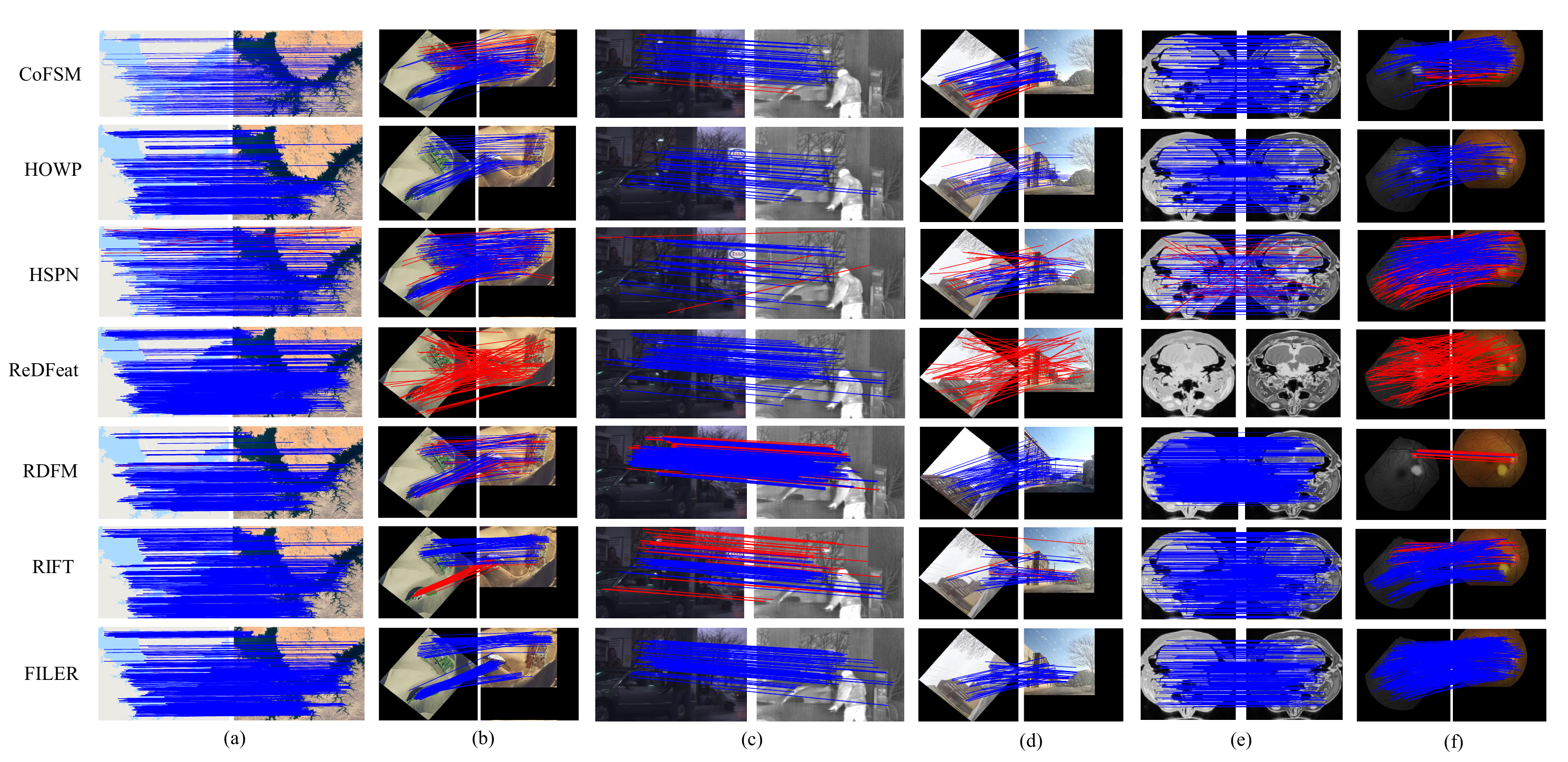}
	\caption{Qualitative results of image matching (The blue lines are true matches, and the red lines are fasle matches). (a) and (b) are map-optical and optical cross-temporal image pairs from remote sensing dataset, respectively. (c) and (d) are visible-IR and visible cross-season image pairs from computer vision dataset, respectively. (e) and (f) are PD-T2 and retinal image pairs from the medical dataset, respectively.}\label{fig8}
\end{figure*}
\begin{figure*}[h]	
	\centering	
	\includegraphics[scale=0.39]{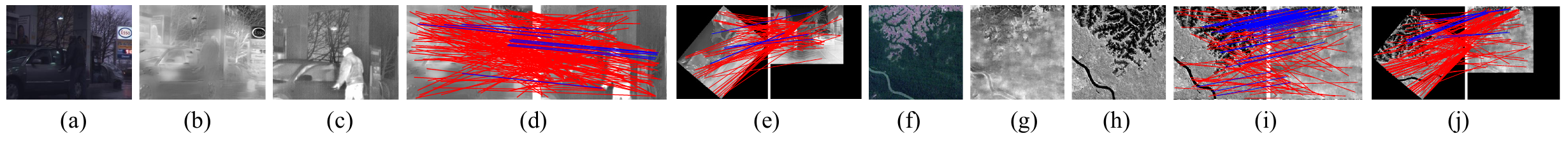}
	\caption{Qualitative results of CPSTN+SIFT (The blue lines are true matches, and the red lines are fasle matches). (a) and (f) are visible and optical images, respectively; (b) and (g) are pseudo infrared images; (c) and (h) are infrared images; (d) and (j) are image matching line diagrams; (e) and (j) are image matching line diagrams after rotating $50^{\circ}$.}\label{fig9}
\end{figure*}	
\subsection{Performance Testing of Feature Matching}
To test the feature matching performance of FILER, we compare it with CoFSM, HOWP, HSPN, ReDFeat, RDFM, RIFT and CPSTN+SIFT in NCM, CMR and runtime. 
\subsubsection{Qualitative Results}
We perform qualitative experiments in two typical image pairs from each category, one of which has the rotation transformation. The results are shown in Figure~\ref{fig8}, where (b), (d) and (f) have rotation of $50^{\circ}$. 

ReDFeat has high NCM and CMR in map-optical and visible-IR, but fails to match PD-T2. ReDFeat is sensitive to rotation and perspective transformation, so the results of rotated images are poor. HSPN has high NCM in map-optical, optical cross-temporal and retinal image pairs, but the number of false matches is also high and hence the precision is low. The number of matches of RDFM and RIFT is high, but the CMR is low, and the results of RDFM in PD-T2 are failed. HOWP has high CMR but low NCM. We use CPSTN to convert visible images into pseudo infrared images and establish matches between infrared and pseudo infrared images using SIFT. In Figure~\ref{fig9} (d) and (j), although there are lots of matches in both image pairs, the accuracy is low. It is mainly because the universality of CPSTN is poor. In Figure~\ref{fig9}, there are still NIDs between the pseudo infrared and infrared images, and the edge structures are not clearer. This leads to low accuracy in feature point positioning and matching. CoFSM and HOWP achieve good matching results in all image pairs, but they still have some gaps compared to FILER.
\subsubsection{Quantitative Results}
\begin{figure*}[t]	
	\centering	
	\includegraphics[scale=0.18]{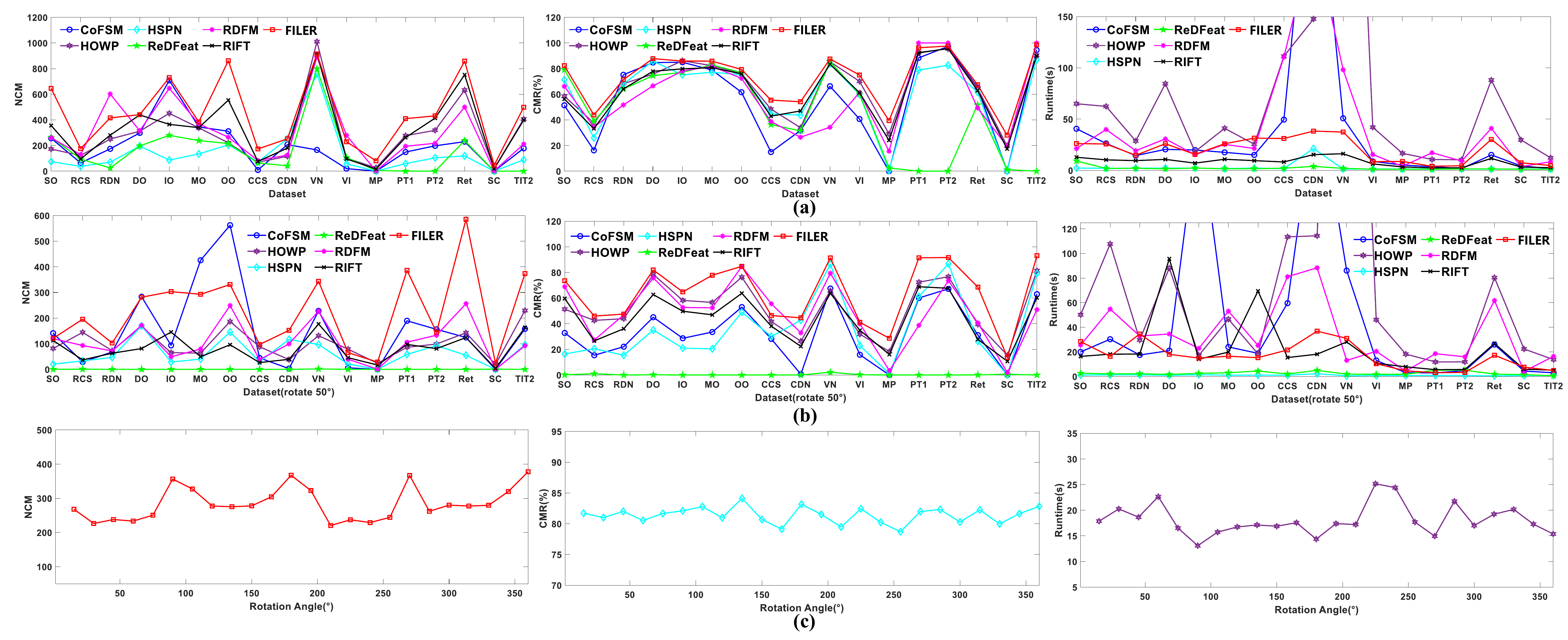}
	\caption{Quantitative results of image matching and angle robustness test. The images in (a) have no rotation transformation. The images in (b) are rotated by $50^{\circ}$. (c) is the result of angle robustness test, whose abscissa is rotation angle $\theta = [0 ^ {\circ}, 15 ^ {\circ}, 30 ^ {\circ},..., 360 ^ {\circ}] $.}\label{fig11}
\end{figure*}

We conduct quantitative experiments on three datasets, consisting of 17 types and 164 pairs of multimodal images. Figure~\ref{fig11} shows the visual line charts of NCM, CMR and runtime of all algorithms, respectively. The abscissa represents the types of image pairs, which are SAR-optical (SO), remote sensing cross-season (RCS), remote sensing day-night (RDN), depth-optical (DO), IR-optical (IO), map-optical (MO), optical-optical (OO), computer vision cross-season (CCS), computer vision day-night (CDN), visible-NIR (VN), visible-IR (VI), MRI-PET (MP), PD-T1 (PT1), PD-T2 (PT2), retinal (Ret), SPECT-CT (SC) and MRI T1-T2 (T1T2) image pairs from left to right, respectively. The image pairs of type 1 to 7, 8 to 11 and 12 to 17 are from remote sensing, computer vision, and medical datasets, respectively. In Figure~\ref{fig11} (a) and Table~\ref{table7}, CoFSM has highest CMR in RDN, and $CMR=75.18\%$. It also has high NCM in some image pairs. CoFSM uses the CoF algorithm to construct an image scale space, and uses the optimized image gradients to extract feature points in the new scale space, which increases the number of feature points. In VN, the NCM of CoFSM decreases significantly, mainly because the information in VN is more complex and chaotic. The NCM and CMR of CoFSM on MP and SC are 0. It is because the information in MP and SC is so different from each other that it is difficult to obtain clues even with manual matching. In addition, the runtime of CoFSM is generally high, especially in CDN and VN. The texture information of them is complex and discontinuous, which leads to the complex computation in the construction of multi-layer CoF scale space.
\begin{table*}[t]
	\fontsize{7pt}{7pt}\selectfont
	\renewcommand\arraystretch{1}   
	\centering
	\caption{NCM and CMR of CoFSM, HOWP, HSPN, ReDFeat, RDFM, RIFT, CPSTN+SIFT, and FILER (The red fonts are the maximum).}
	\label{table7}
	
	\setlength{\tabcolsep}{0.2mm}{
		\begin{tabular}{|c|c|c|c|c|c|c|c|c|c|c|c|c|c|c|c|c|c|c|}			
			\hline
			\rule{0pt}{7pt} 			
			\multirow{2}{*}{Method}&\multirow{2}{*}{Index}& \multicolumn{7}{c}{Remote sensing}& \multicolumn{4}{|c}{Computer vision} &\multicolumn{6}{|c|}{Medical}\\ 
			\cline{3-9}  \cline{10-13} \cline{14-19} &&SO & RCS & RDN & DO & IO & MO&OO & CCS & CDN & VN & VI  & MP & PT1 & PT2 & Ret & SC & T1T2 \\			
			\hline
			\rule{0pt}{12pt} 
			CoFSM \cite{yao2022multi} &\makecell[c]{NCM \\ CMR($\%$)}	&\makecell[c]{256.3\\51.38}	&\makecell[c]{64.20\\ 16.24}	&\makecell[c]{174.3 \\ \textcolor{red}{\textbf{75.18}}}	&\makecell[c]{298.4 \\84.71}	&\makecell[c]{708.3 \\ 84.97}	&\makecell[c]{344.9\\ 78.94}	&\makecell[c]{311.5\\ 61.61}&\makecell[c]{9.167 \\14.92}& \makecell[c]{207.0\\ 31.73}&\makecell[c]{165.5\\ 66.25}
			&\makecell[c]{19.73\\40.68}&\makecell[c]{0 \\0}&\makecell[c]{149.0\\88.34}&\makecell[c]{197.6\\ 97.27}&\makecell[c]{230.2 \\65.29}&\makecell[c]{0 \\0}
			&\makecell[c]{178.2 \\94.46}	\\ 
			\hline
			\rule{0pt}{12pt} 			
			HOWP \cite{zhang2023histogram} &\makecell[c]{NCM \\ CMR($\%$)}	&\makecell[c]{172.5\\58.44}	&\makecell[c]{122.0\\ 16.24}	&\makecell[c]{250.5\\68.23}	&\makecell[c]{311.8 \\75.59}	&\makecell[c]{451.0\\ \textcolor{red}{\textbf{86.12}}}	&\makecell[c]{338.0\\ 82.29}	&\makecell[c]{221.3\\ 76.23}&\makecell[c]{69.43 \\48.40}& \makecell[c]{116.0\\33.91}&\makecell[c]{\textcolor{red}{\textbf{1011}}\\ 85.40}
			&\makecell[c]{101.7\\70.11}&\makecell[c]{26.90\\29.02}&\makecell[c]{276.8\\92.61}&\makecell[c]{318.6\\ 95.44}&\makecell[c]{632.5 \\64.83}&\makecell[c]{19.27 \\19.77}
			&\makecell[c]{406.9 \\90.45}	\\ 
			\hline
			\rule{0pt}{12pt} 			
			HSPN \cite{zhao2022heterogeneous}&\makecell[c]{NCM \\ CMR($\%$) }	&\makecell[c]{74.50\\71.28}	&\makecell[c]{37.40 \\25.98}	&\makecell[c]{70.00\\ 68.00}	&\makecell[c]{195.8\\85.85}	&\makecell[c]{86.50\\ 75.15}	&\makecell[c]{134.3\\77.13}	&\makecell[c]{205.2\\ 75.35}&\makecell[c]{71.67\\ 44.69}& \makecell[c]{252.7\\ 43.71}&\makecell[c]{759.7\\83.74}&\makecell[c]{57.55\\59.75}				
			&\makecell[c]{0.100 \\0.625}&\makecell[c]{59.30\\78.83}&\makecell[c]{104.1\\ 82.62}&\makecell[c]{119.0\\61.59}&\makecell[c]{0 \\0}&\makecell[c]{88.70\\86.82}	\\ 
			\hline
			\rule{0pt}{12pt}				
			ReDFeat \cite{deng2022redfeat}&\makecell[c]{NCM \\ CMR($\%$) }	&\makecell[c]{264.8\\ 79.18}	&\makecell[c]{94.40 \\39.09}	&\makecell[c]{24.75\\ 64.08}	&\makecell[c]{197.0 \\74.48}	&\makecell[c]{279.5\\ 77.08}	&\makecell[c]{238.6 \\82.93}	&\makecell[c]{215.3\\ 76.83}&\makecell[c]{64.17\\ 36.34}& \makecell[c]{41.67\\ 31.42}&\makecell[c]{800.3\\85.38}
			&\makecell[c]{106.4 \\60.23}&\makecell[c]{5.9\\2.510}&\makecell[c]{0\\ 0}&\makecell[c]{0\\0}&\makecell[c]{238.6 \\51.72}&\makecell[c]{0.200\\1.204}
			&\makecell[c]{0 \\0}	\\ 
			\hline
			\rule{0pt}{12pt}
			RDFM \cite{cao2023rdfm}&\makecell[c]{NCM \\ CMR($\%$) }	&\makecell[c]{260.5\\ 66.19}	&\makecell[c]{129.2 \\34.37}	&\makecell[c]{\textcolor{red}{\textbf{602.0}}\\ 51.66}	&\makecell[c]{315.1 \\66.51}	&\makecell[c]{646.5\\ 78.14}	&\makecell[c]{368.7 \\81.79}	&\makecell[c]{265.2\\ 72.40}&\makecell[c]{83.29\\ 38.73}& \makecell[c]{125.7\\ 26.48}&\makecell[c]{891.0\\34.25}
			&\makecell[c]{\textcolor{red}{\textbf{279.4}} \\61.45}&\makecell[c]{2.300\\15.58}&\makecell[c]{194.0\\ \textcolor{red}{\textbf{100.0}}}&\makecell[c]{217.6\\\textcolor{red}{\textbf{100.0}}}&\makecell[c]{499.7 \\49.31}&\makecell[c]{2.100\\19.48}
			&\makecell[c]{211.3 \\\textcolor{red}{\textbf{100.0}}}	\\ 
			\hline
			\rule{0pt}{12pt}  			
			RIFT \cite{li2019rift}&\makecell[c]{NCM \\ CMR($\%$) }	&\makecell[c]{356.2\\ 56.27}	&\makecell[c]{95.80 \\33.22}	&\makecell[c]{280.8\\ 64.19}	&\makecell[c]{438.6 \\77.91}	&\makecell[c]{365.0\\ 79.80}	&\makecell[c]{339.4 \\80.76}	&\makecell[c]{553.8\\ 75.91}&\makecell[c]{75.83\\ 42.92}& \makecell[c]{180.0\\ 46.99}&\makecell[c]{914.9\\ 83.45}
			&\makecell[c]{95.55 \\61.24}&\makecell[c]{15.80\\ 24.08}&\makecell[c]{262.6\\ 92.21}&\makecell[c]{413.8\\95.74}&\makecell[c]{751.7 \\62.86}&\makecell[c]{18.70\\ 17.39}
			&\makecell[c]{399.0 \\89.83}	\\ 
			\hline
			\rule{0pt}{12pt} 
			\makecell[c]{CPSTN \\ +SIFT} \cite{wang2022unsupervised} &\makecell[c]{NCM \\ CMR($\%$)}	&\makecell[c]{\textendash}	&\makecell[c]{\textendash}	&\makecell[c]{\textendash}	&\makecell[c]{\textendash}&\makecell[c]{2.750\\0.565}&\makecell[c]{\textendash}	&\makecell[c]{\textendash}	&\makecell[c]{\textendash}	&\makecell[c]{\textendash}&\makecell[c]{\textendash}& \makecell[c]{8.182\\ 7.819}&\makecell[c]{\textendash}
			&\makecell[c]{\textendash}&\makecell[c]{\textendash}&\makecell[c]{\textendash}&\makecell[c]{\textendash}
			&\makecell[c]{\textendash}	\\ 
			\hline
			\rule{0pt}{12pt} 
			FILER &\makecell[c]{NCM \\ CMR($\%$)}	&\makecell[c]{\textcolor{red}{\textbf{645.8}}\\ \textcolor{red}{\textbf{82.28}}}	&\makecell[c]{\textcolor{red}{\textbf{176.0}}\\ \textcolor{red}{\textbf{43.84}}}	&\makecell[c]{415.8\\ 71.46}	&\makecell[c]{\textcolor{red}{\textbf{440.4}}\\ \textcolor{red}{\textbf{87.77}}}	&\makecell[c]{\textcolor{red}{\textbf{730.3}}\\ 85.91}	&\makecell[c]{\textcolor{red}{\textbf{384.3}}\\ \textcolor{red}{\textbf{85.83}}}	&\makecell[c]{\textcolor{red}{\textbf{861.8}}\\ \textcolor{red}{\textbf{79.38}}}&\makecell[c]{\textcolor{red}{\textbf{173.1}}\\ \textcolor{red}{\textbf{55.44}}}& \makecell[c]{\textcolor{red}{\textbf{252.3}}\\ \textcolor{red}{\textbf{54.14}}}&\makecell[c]{914.7\\ \textcolor{red}{\textbf{87.51}}}
			&\makecell[c]{229.3\\ \textcolor{red}{\textbf{75.04}}}&\makecell[c]{\textcolor{red}{\textbf{80.80}}\\ \textcolor{red}{\textbf{39.45}}}&\makecell[c]{\textcolor{red}{\textbf{409.8}}\\ 96.28}&\makecell[c]{\textcolor{red}{\textbf{431.4}} \\97.61 }&\makecell[c]{\textcolor{red}{\textbf{858.2}}\\ \textcolor{red}{\textbf{67.43}}}&\makecell[c]{\textcolor{red}{\textbf{46.60}}\\ \textcolor{red}{\textbf{28.10}}}
			&\makecell[c]{\textcolor{red}{\textbf{498.5}}\\ 98.60}	\\  
			\hline			
		\end{tabular}
	}	
\end{table*}

RDFM has the highest NCM in RDN and VI image pairs, and the highest CMR in PT1, PT2 and TIT2 image pairs. Because RDFM can extract the robust deep feature of the images, its NCM is higher, and its NCMs are better than other algorithms except FILER. But the runtime of RDFM is high, especially for CDN and CCS. The overall CMR of HOWP algorithm is higher than that of other algorithms, and it has the largest CMR in IO image pairs and the largest NCM in VN image pairs. It is because HOWP's regularization-based log-polar descriptor improves its robustness to nonlinear radiation distortion and contrast differences. However, the optimization of feature points by the feature aggregation strategy reduces the number of feature points, so the NCM is low. HOWP runtime is also high, especially in CDN and VN. Because RDFM and HOWP are less robust to noises and local nonlinear geometric distortions, the overall performance is lower than FILER.

RIFT successfully matches all images. The CMR of RIFT is better than that of RDFM in computer vision images. RIFT detects feature points by phase congruency, and designs maximum orientation map based on FDF convolution image sequences to construct descriptors. RIFT has good generality for MIM and is more robust to NIDs than the traditional gradient methods. In Figure~\ref{fig11}, the performance of RIFT is better than CoFSM.

In Table~\ref{table7}, the quantitative results of the CPSTN+SIFT method in IO and VI are not good. This is because the NIDs between the pseudo infrared and the infrared images have not been effectively reduced and the common edge structures are less, resulting in poor positioning accuracy of feature points and matching performance.

Table~\ref{table7} and Figure~\ref{fig11} shows the average NCM, CMR and run time of all methods in all image pairs. The results show that FILER has clear advantages over other algorithms in most image pairs, and the run time of FILER is slightly larger than RIFT, but smaller than CoFSM, HOWP and RDFM. Because FILER detects and describes features based on the local energy response model, it largely preserves the common information and details, and also adopts the edge structure enhancement and convolutional feature weighting to suppress noises and local nonlinear geometric distortions. ReDFeat performs well in some image pairs, such as RCS and VI image pair, but poor in PT1, PT2 and TIT2. However, its overall performance is weaker than RIFT and RDFM. The performance of HSPN is weaker than ReDFeat. The overall performance CPSTN+SIFT is significantly worse than the other methods.

\begin{table*}[t]
	\fontsize{7pt}{7pt}\selectfont
	\renewcommand\arraystretch{1}   
	\centering
	\caption{NCM and CMR of CoFSM, HOWP, HSPN, ReDFeat, RDFM, RIFT, CPSTN+SIFT and FILER in image pairs with $50^{\circ}$ rotation (The red fonts are the maximum).}
	\label{table8}
	
	\setlength{\tabcolsep}{0.2mm}{
		\begin{tabular}{|c|c|c|c|c|c|c|c|c|c|c|c|c|c|c|c|c|c|c|}			
			\hline
			\rule{0pt}{7pt} 			
			\multirow{2}{*}{Method}&\multirow{2}{*}{Index}& \multicolumn{7}{c}{Remote sensing}& \multicolumn{4}{|c}{Computer vision} &\multicolumn{6}{|c|}{Medical}\\ 
			\cline{3-9}  \cline{10-13} \cline{14-19} &&SO & RCS & RDN & DO & IO & MO&OO & CCS & CDN & VN & VI  & MP & PT1 & PT2 & Ret & SC & T1T2 \\			
			\hline
			\rule{0pt}{12pt} 
			CoFSM \cite{yao2022multi} &\makecell[c]{NCM \\ CMR($\%$)}	&\makecell[c]{\textcolor{red}{\textbf{141.2}}\\32.82}	&\makecell[c]{29.00\\ 15.43}	&\makecell[c]{67.75 \\22.08}	&\makecell[c]{\textcolor{red}{\textbf{283.4}} \\45.07}	&\makecell[c]{93.50\\ 28.66}	&\makecell[c]{\textcolor{red}{\textbf{425.7}}\\33.53}	&\makecell[c]{\textcolor{red}{\textbf{562.5}}\\52.99}&\makecell[c]{44.50 \\28.16}& \makecell[c]{3.000\\0.615}&\makecell[c]{228.1\\ 67.46}
			&\makecell[c]{6.000\\15.82}&\makecell[c]{0 \\0}&\makecell[c]{189.5\\60.09}&\makecell[c]{\textcolor{red}{\textbf{156.2}}\\ 67.22}&\makecell[c]{124.4 \\31.11}&\makecell[c]{0 \\0}
			&\makecell[c]{158.0 \\63.07}	\\ 
			\hline
			\rule{0pt}{12pt} 				
			HOWP \cite{zhang2023histogram} &\makecell[c]{NCM \\ CMR($\%$)}	&\makecell[c]{81.50\\51.33}	&\makecell[c]{144.4\\ 42.62}	&\makecell[c]{73.00 \\43.96}	&\makecell[c]{166.9 \\78.48}	&\makecell[c]{65.25\\ 58.23}	&\makecell[c]{63.43\\56.63}	&\makecell[c]{186.7\\76.33}&\makecell[c]{88.17 \\41.71}& \makecell[c]{38.00\\26.39}&\makecell[c]{131.3\\ 65.78}
			&\makecell[c]{\textcolor{red}{\textbf{78.82}}\\31.83}&\makecell[c]{23.60 \\18.37}&\makecell[c]{86.20\\72.38}&\makecell[c]{98.90\\76.72}&\makecell[c]{142.6 \\39.54}&\makecell[c]{19.20 \\15.41}
			&\makecell[c]{229.6 \\81.42}	\\ 
			\hline
			\rule{0pt}{12pt} 				
			HSPN \cite{zhao2022heterogeneous}&\makecell[c]{NCM \\ CMR($\%$) }	&\makecell[c]{20.33\\16.41}	&\makecell[c]{33.00\\20.51}	&\makecell[c]{46.75\\ 15.41}	&\makecell[c]{169.6 \\35.11}	&\makecell[c]{28.50\\ 20.96}	&\makecell[c]{40.00 \\20.39}	&\makecell[c]{145.7\\49.07}&\makecell[c]{30.17\\ 30.06}& \makecell[c]{117.0\\ 42.74}&\makecell[c]{97.07\\ 86.39}
			&\makecell[c]{20.91 \\22.94}	
			&\makecell[c]{0.400\\ 0.920}&\makecell[c]{58.60\\61.24}&\makecell[c]{92.50\\86.72}&\makecell[c]{54.96\\26.28}&\makecell[c]{0\\ 0}
			&\makecell[c]{96.00 \\79.16}	\\			
			\hline
			\rule{0pt}{12pt}				
			ReDFeat \cite{deng2022redfeat}&\makecell[c]{NCM \\ CMR($\%$) }	&\makecell[c]{0.167\\0.165}	&\makecell[c]{0.800\\1.006}	&\makecell[c]{0\\ 0}	&\makecell[c]{0.143 \\0.210}	&\makecell[c]{0\\ 0}	&\makecell[c]{0 \\0}&\makecell[c]{0 \\0}	
			&\makecell[c]{0\\0}&\makecell[c]{0\\ 0}& \makecell[c]{1.700\\2.168}&\makecell[c]{0.182\\ 0.233}	
			&\makecell[c]{0 \\0}&\makecell[c]{0\\ 0}&\makecell[c]{0\\0}&\makecell[c]{0.174\\0.132}&\makecell[c]{0.700 \\0.557}&\makecell[c]{0\\ 0}	\\
			\hline
			\rule{0pt}{12pt} 			
			RDFM \cite{cao2023rdfm}&\makecell[c]{NCM \\ CMR($\%$) }	&\makecell[c]{119.8\\68.89}	&\makecell[c]{93.20\\27.27}	&\makecell[c]{72.00\\ 46.93}	&\makecell[c]{171.9 \\75.72}	&\makecell[c]{49.00\\ 52.69}	&\makecell[c]{79.43 \\52.44}&\makecell[c]{249.0 \\84.69}	
			&\makecell[c]{38.00\\55.57}&\makecell[c]{99.33\\ 32.89}	
			&\makecell[c]{227.1 \\79.42}&\makecell[c]{39.91\\ 39.89}&\makecell[c]{0.600\\3.393}&\makecell[c]{106.2\\38.74}&\makecell[c]{133.3 \\73.62}&\makecell[c]{256.2\\ 40.63}&\makecell[c]{0.200\\ 1.818}&\makecell[c]{92.20\\ 51.05}	\\
			\hline
			\rule{0pt}{12pt} 	
			RIFT \cite{li2019rift}&\makecell[c]{NCM \\ CMR($\%$) }	&\makecell[c]{112.5\\59.68}	&\makecell[c]{37.00\\26.65}	&\makecell[c]{63.00\\ 36.09}	&\makecell[c]{81.25 \\62.74}	&\makecell[c]{145.5\\ 49.75}	&\makecell[c]{49.43 \\46.96}	&\makecell[c]{96.00\\63.76}&\makecell[c]{26.50\\ 38.07}& \makecell[c]{39.33\\ 22.35}&\makecell[c]{176.9\\ 63.73}
			&\makecell[c]{45.60 \\34.91}&\makecell[c]{17.00\\ 16.05}&\makecell[c]{97.30\\68.86}&\makecell[c]{80.90\\67.69}&\makecell[c]{124.1 \\27.79}&\makecell[c]{3.600\\ 10.68}
			&\makecell[c]{163.5 \\60.30}	\\
			\hline
			\rule{0pt}{12pt} 
			\makecell[c]{CPSTN \\+SIFT} \cite{wang2022unsupervised} &\makecell[c]{NCM \\ CMR($\%$)}	&\makecell[c]{\textendash}	&\makecell[c]{\textendash}	&\makecell[c]{\textendash}	&\makecell[c]{\textendash}&\makecell[c]{2.000\\0.321}&\makecell[c]{\textendash}	&\makecell[c]{\textendash}	&\makecell[c]{\textendash}	&\makecell[c]{\textendash}& \makecell[c]{\textendash}
			&\makecell[c]{3.818\\1.601}&\makecell[c]{\textendash}&\makecell[c]{\textendash}&\makecell[c]{\textendash}
			&\makecell[c]{\textendash}&\makecell[c]{\textendash} &\makecell[c]{\textendash}	\\ 				 
			\hline
			\rule{0pt}{12pt} 				
			FILER &\makecell[c]{NCM \\ CMR($\%$)}	&\makecell[c]{121.0\\ \textcolor{red}{\textbf{73.56}}}	&\makecell[c]{\textcolor{red}{\textbf{195.6}}\\ \textcolor{red}{\textbf{45.98}}}	&\makecell[c]{\textcolor{red}{\textbf{102.8}}\\ \textcolor{red}{\textbf{47.53}}}	&\makecell[c]{281.6\\ \textcolor{red}{\textbf{81.96}}}	&\makecell[c]{\textcolor{red}{\textbf{303.3}}\\ \textcolor{red}{\textbf{64.84}}}	&\makecell[c]{292.9\\\textcolor{red}{\textbf{77.85}}}	&\makecell[c]{331.0\\ \textcolor{red}{\textbf{84.79}}}
			&\makecell[c]{\textcolor{red}{\textbf{97.00}}\\ \textcolor{red}{\textbf{46.42}}}& \makecell[c]{\textcolor{red}{\textbf{152.3}}\\ \textcolor{red}{\textbf{44.60}}}&\makecell[c]{\textcolor{red}{\textbf{343.3}}\\ \textcolor{red}{\textbf{91.41}}}
			&\makecell[c]{65.27\\ \textcolor{red}{\textbf{41.10}}}&\makecell[c]{\textcolor{red}{\textbf{28.50}}\\ \textcolor{red}{\textbf{28.65}}}&\makecell[c]{\textcolor{red}{\textbf{386.4}}\\ \textcolor{red}{\textbf{91.45}}}&\makecell[c]{144.7 \\ \textcolor{red}{\textbf{91.65}} }&\makecell[c]{\textcolor{red}{\textbf{585.3}}\\ \textcolor{red}{\textbf{68.57}}}&\makecell[c]{\textcolor{red}{\textbf{23.70}}\\ \textcolor{red}{\textbf{13.62}}}
			&\makecell[c]{\textcolor{red}{\textbf{374.0}}\\\textcolor{red}{\textbf{93.15}}}	\\  
			\hline			
		\end{tabular}
	}	
\end{table*}
\subsection{Rotation Invariance Test}
FILER achieves rotation invariance and compares with other algorithms.
\subsubsection{Comparison Experiment}
Figure~\ref{fig8} shows the qualitative results on the multimodal images rotated by $50^{\circ}$ and FILER achieves the best results. Figure~\ref{fig11} (b) and Table~\ref{table8} show the quantitative results after rotating $50^{\circ}$. In Figure~\ref{fig11} (b), the NCM and CMR of all algorithms decrease to some extent. CoFSM still maintains the advantage of NCM in SO, DO, MO, OO and PT2, but FILER is outperformed on the remaining image pairs. The performance of RIFT degrades significantly because its angle-traversing strategy takes only six angles in $[0^{\circ}, 180^{\circ}]$. So when the image is rotated by $50^{\circ}$, the closest angle it traverses is $30^{\circ}$, which has an error of $20^{\circ}$. Moreover, it can only handle images with rotation angles in $[0^{\circ}, 180^{\circ}]$. If the angle exceeds this range, the performance deteriorates even more. ReDFeat is sensitive to rotation and perspective transformation, so its performance is poor. HSPN uses a homologous matrix to establish correspondences between images, which can cope with rotation transformation, but the performance is still degraded. HOWP only maintains the advantage of NCM in VI. The performance of HOWP and RDFM is significantly reduced when the rotation angle is large. Our method has little change in CMR and NCM, and still has advantages. CoFSM, HOWP and RDFM is still very time-consuming, but FILER has almost the same runtime as RIFT for all image pairs except CDN. Qualitative and quantitative results show that FILER is more robust to rotation changes than other algorithms.
\begin{figure*}[t]	
	\centering	
	\includegraphics[scale=0.3]{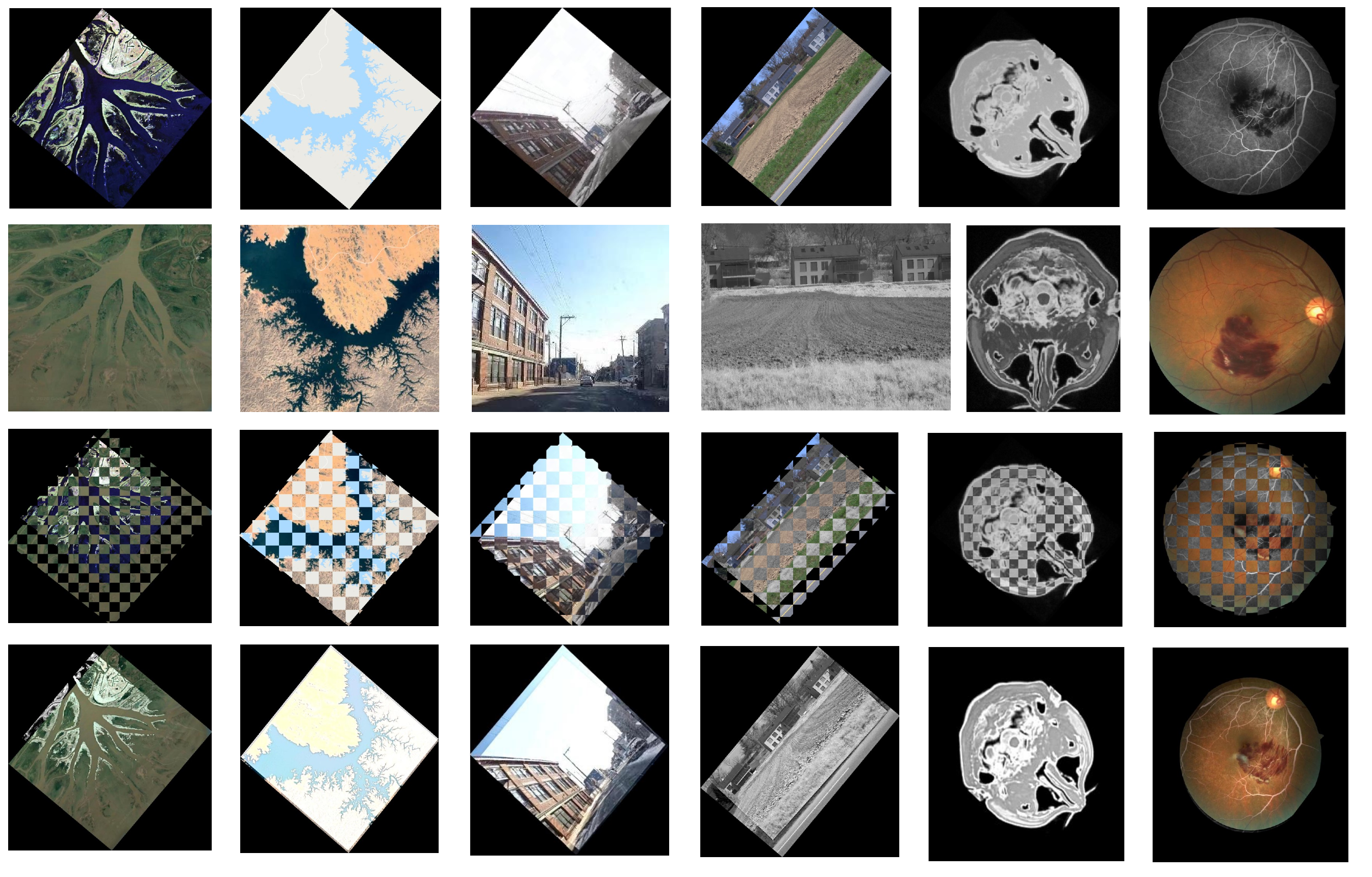}
	\caption{The results of image registration and fusion. The first and second rows are multimodal image pairs. The images in the first row are rotated by $50^{\circ}$. The third row is the result of checkerboard mosaic, and the fourth row is the result of image fusion.}\label{fig12}
\end{figure*}	
\subsubsection{Angle Robustness Test}
To test the robustness of FILER for different angles, we uniformly select 24 angles in $[0^{\circ}, 360^{\circ}]$ separated by $15^{\circ}$. We use FILER to match the image pairs rotated by $\theta^\circ$, and record the average NCM, CMR and runtime at each angle. In Figure~\ref{fig11} (c), the abscissa is the rotation angle, and the ordinate are NCM, CMR and runtime, respectively. Although the NCM, CMR, and runtime are different for each angle, their values generally vary within a small range. The NCMs and CMRs at all angles are greater than 200 and 80$\%$, respectively, and runtimes are lower than 25s, which shows that FILER is robust to the rotation in $[0^{\circ}, 360^{\circ}]$.
\subsection{Image Registration and Fusion Test}
In order to test the performance of FILER in practical applications, we conduct image registration and fusion experiments. We use SwinFusion \cite{ma2022swinfusion} network based on Swin Transformer to conduct image fusion.	

First, we select two typical image pairs from each category for registration and fusion. Each image pair is rotated by $50^\circ$. In Figure~\ref{fig12}, the registration and fusion results are good in all image pairs, with essentially no misalignment and artifacts, which proves that the matches generated by FILER have high accuracy and reasonable distribution.

\begin{table}[t]
	\fontsize{10pt}{10pt}\selectfont
	\renewcommand\arraystretch{1}   
	\centering
	\caption{RMSE, ME and SR of all algorithms (The red fonts are the minimum).}
	\label{table9}
	\setlength{\tabcolsep}{1.8mm}{
		\begin{tabular}{|c|c|c|c|c|c|c|c|c|}		
			\hline
			\rule{0pt}{12pt}			
			Dataset&Index&\makecell[c]{CoFSM \\ \cite{yao2022multi}}& \makecell[c]{HOWP\\  \cite{zhang2023histogram}} & \makecell[c]{HSPN \\ \cite{zhao2022heterogeneous}}&\makecell[c]{ReDFeat\\ \cite{deng2022redfeat}}&\makecell[c]{RDFM\\ \cite{cao2023rdfm}}&\makecell[c]{RIFT\\ \cite{li2019rift}} &FILER\\
			
			\hline	
			\rule{0pt}{15pt} 
			
			\makecell[c]{Remote\\sensing}& \makecell[c]{RMSE \\ME\\ SR} &\makecell[c]{42.98\\ 42.15\\ 0.900}
			&\makecell[c]{9.192\\ 8.124\\0.950} &\makecell[c]{20.56\\13.87\\0.897}
			&\makecell[c]{107.1\\46.12\\1.000}  &\makecell[c]{11.35\\ 10.64\\0.750}
			&\makecell[c]{3.403\\ 3.013\\1.000} &\makecell[c]{\textcolor{red}{\textbf{2.518}}\\ \textcolor{red}{\textbf{2.220}}\\ \textcolor{red}{\textbf{1.000}}}\\

			\hline
			\rule{0pt}{15pt} 
			
			\makecell[c]{Computer\\vision}&\makecell[c]{RMSE \\ME\\ SR} &\makecell[c]{34.23\\33.49\\0.745}&\makecell[c]{7.652\\2.962\\0.961}
			&\makecell[c]{26.20\\21.69\\0.940}&\makecell[c]{113.2\\52.14\\0.961}
			&\makecell[c]{10.41\\9.219\\0.857}
			&\makecell[c]{6.792\\4.278\\0.961}
			&\makecell[c]{\textcolor{red}{\textbf{4.084}}\\ \textcolor{red}{\textbf{3.167}}\\ \textcolor{red}{\textbf{0.961}}}\\
			
			\hline
			\rule{0pt}{15pt} 
			Medical&\makecell[c]{RMSE \\ME\\ SR} 
			&\makecell[c]{9.724 \\9.173\\ 0.959}&\makecell[c]{4.289\\ 3.805\\ 0.959}
			&\makecell[c]{19.42\\ 16.78\\ 0.698}&\makecell[c]{86.39\\46.89\\ 0.744}&\makecell[c]{7.926\\6.861\\0.850}
			&\makecell[c]{4.419\\3.855\\0.986}
			&\makecell[c]{\textcolor{red}{\textbf{3.857}}\\ \textcolor{red}{\textbf{ 3.458}}\\ \textcolor{red}{\textbf{1.000}}}\\
			\hline			
		\end{tabular}
	}	
\end{table}
Subsequently, we compare the average root mean square error (RMSE), mean error (ME) and success rate (SR) \cite{li2019rift,yao2022multi} of all algorithms. We compute the RMSE and ME using the affine transformation matrix of groundtruth. SR represents the proportion of image pairs with $NCM>4$ in data set. Table~\ref{table9} shows the quantitative results, and HOWP, RIFT and FILER have advantages. Both the RMSE and ME of FILER are within 3 pixels, and the SR is above 0.95. The overall performance of FILER is better than RIFT and HOWP. The CMR can influence the accuracy of the estimated transformation model, and a low CMR leads to large location error of the feature points, which results in large RMSE and ME. Although RDFM and CoFSM has higher NCM in some images, the CMR and SR are relatively low, which makes RMSE and ME relatively large. The RSME and ME of ReDFeat and HSPN are large due to the low CMR.

\section{Conclusion}
To overcome the NIDs of MIM, a novel multimodal feature matching method FILER is proposed. The core of FILER is local energy response model, which is also the basis of feature detector and descriptor. In feature detection, the edge structure enhanced feature detector is constructed by total energy response map, which can detect more feature points with high repeatability and improve the robustness to noises. In feature description, the energy manifold vector field is obtained from convolutional feature weighted local energy response layers, and log-polar descriptor is constructed based on it, which improve the robustness to local nonlinear geometric distortions. In addition, we also design the main direction method to achieve rotation invariance.

FILER achieves good matching in all image pairs from data set. The results of feature detection, feature matching, and rotation invariance tests are significantly better than CoFSM, HOWP, HSPN, ReDFeat, RDFM, RIFT and CPSTN+SIFT. FILER also performs well in image registration and fusion. The reliability and superiority of FILER are verified by qualitative and quantitative comparisons. In addition, the local energy response model of the FILER can be designed into trainable network structures that can be combined with deep learning methods such as Convolutional Neural Network (CNN) and Generative Adversarial Network (GAN) to improve future matching performance.

\section*{Acknowledgment}
This work was supported by the National Natural Science Foundation of China nos. 62073304 and 62373338.




\bibliographystyle{elsarticle-num} 
\bibliography{refsd.bib}


%
%

\end{document}